\documentclass{article}
\usepackage[preprint,nonatbib]{neurips_2020}
\usepackage{graphicx}
\usepackage{comment}
\usepackage{amsmath,amssymb,amsfonts} %
\usepackage{color}

\usepackage{mathtools}
\usepackage{nicefrac}
\usepackage{url}
\usepackage[dvipsnames]{xcolor}
\usepackage{wrapfig}
\usepackage{titlesec}
\usepackage{multirow}
\usepackage{subfig}
\usepackage{enumitem}
\usepackage[normalem]{ulem}
\useunder{\uline}{\ul}{}

\usepackage{adjustbox}

\usepackage[pagebackref=true,breaklinks=true,colorlinks,bookmarks=false]{hyperref}
\usepackage{cleveref}
\usepackage{tikz}
\usetikzlibrary{arrows}

\titlespacing*{\section}{0pt}{.1ex plus .0ex minus .0ex}{.1ex plus .0ex minus .0ex}
\titlespacing*{\subsection}{0pt}{-.0ex plus .0ex minus .0ex}{-.0ex plus .0ex minus .0ex}
\titlespacing*{\paragraph}{0pt}{-.0ex}{1em}
\usepackage[font=small,aboveskip=10pt, belowskip=-10pt]{caption}

\newcommand{\C}{\mathbf{C}}
\newcommand{\X}{\mathbf{X}}
\newcommand{\x}{\mathbf{x}}
\newcommand{\w}{\mathbf{w}}
\newcommand{\K}{\mathbf{K}}
\newcommand{\R}{\mathbf{R}}
\newcommand{\Rt}{\mathbf{R_t}}
\renewcommand{\P}{\mathbf{P}}
\renewcommand{\t}{\mathbf{t}}
\newcommand{\bbR}{\mathbb{R}}

\newcommand{\object}{\mathbb{O}}
\newcommand{\M}{\mathbf{M}}

\newcommand{\I}{\mathbf{I}}
\newcommand{\feature}{\mathcal{F}}

\newcommand{\V}{\mathbf{V}}
\renewcommand{\P}{\mathbf{P}}
\newcommand{\D}{\mathbf{D}}
\newcommand{\depthSet}{\mathcal{D}}

\newcommand{\diag}{\mathrm{diag}}

\begin{document}

\setlength{\abovedisplayskip}{3pt}
\setlength{\belowdisplayskip}{3pt}

\title{Learning to Detect 3D Reflection Symmetry \\ for Single-View Reconstruction}
\author{%
  Yichao Zhou\thanks{This work is supported by a research grant from Sony Research.} \\
  Univ. of California, Berkeley \\
  \texttt{zyc@berkeley.edu} \\
  \And
  Shichen Liu \\
  Univ. of Southern California \\
  \texttt{liushich@usc.edu} \\
  \And
  Yi Ma \\
  Univ. of California, Berkeley \\
  \texttt{yima@eecs.berkeley.edu} \\
}

\maketitle
\vspace{-10pt}
\begin{abstract}
3D reconstruction from a single RGB image is a challenging problem in computer vision. 
Previous methods are usually solely data-driven, which lead to inaccurate 3D shape recovery and limited generalization capability.
In this work, we focus on object-level 3D reconstruction and present a \emph{geometry-based} end-to-end deep learning framework that first detects the mirror plane of reflection symmetry that commonly exists in man-made objects and then predicts depth maps by finding the intra-image pixel-wise correspondence of the symmetry.
Our method fully utilizes the geometric cues from symmetry during the test time by building plane-sweep cost volumes, a powerful tool that has been used in multi-view stereopsis.
To our knowledge, this is the first work that uses the concept of cost volumes in the setting of single-image 3D reconstruction.
We conduct extensive experiments on the ShapeNet dataset and find that our reconstruction method significantly outperforms the previous state-of-the-art single-view 3D reconstruction networks in term of the accuracy of camera poses and depth maps, without requiring objects being completely symmetric.
Code is available at \url{https://github.com/zhou13/symmetrynet}.

\end{abstract}

\section{Introduction}
3D reconstruction is one of the most long lasting problems in computer vision.
Although commercially available solutions using time-of-flight cameras or structured lights may be able to meet the needs of specific purposes, they are often expensive, have limited range, and can be interfered by other light sources.
On the other hand, traditional image-based 3D reconstruction methods, such as structure-from-motion (SfM), visual SLAM, and multi-view stereopsis (MVS) use only RGB images and recover the underlying 3D geometry \cite{hartley2003multiple,MaY2003}.

Recent advances in convolutional neural networks (CNNs) have shown good potential in inferring dense 3D depth map from RGB images by leveraging  supervised learning.
Deep learning-based multi-view stereo and optical flow methods employ 3D CNNs to regularize the cost volumes that are built upon geometric structures to infer depth maps and have shown the state-of-the-art results on various benchmarks \cite{MultiView-ji2017surfacenet,yao2018mvsnet,MultiView-galliani2016just,MultiView-luo2019p,MultiView-xue2019mvscrf,MultiView-chen2019point}.
Due to the popularity of mobile apps, 3D inference from a single image starts to draw increasing attention.
Nowadays, learning-based single-view reconstruction methods are able to predict 3D shapes in various representations with an encoder-decoder CNN that extrapolates objects from seen patterns during the test time.
However, unlike multi-view stereopsis, previous single-view methods hardly exploit the geometric constraints between the input RGB image and the resulting 3D shape.
Hence, the formulation is ill-posed, and leads to inaccurate 3D shape recovery and limited generalization capability \cite{Equivalent2ImageClassification}.

To address the illness, we identify a structure that commonly exists in man-made objects, the \emph{reflection symmetry}, as a geometric connection between the depth maps and the images, and incorporate it as a prior into deep networks through plane-sweep cost volumes built from features of corresponding pixels, aiming to faithfully recover 3D shapes from single-view images under the principle of shape-from-symmetry~\cite{HongW2004}.
To this end, we propose \emph{SymmetryNet} that combines the strength of learning-based recognition and geometry-based reconstruction methods.
Specifically, SymmetryNet first detects the parameters of the mirror plane from the image with a coarse-to-fine strategy and then recovers the depth from a reflective stereopsis.
The network consists of a backbone feature extractor, a differentiable warping module for building the 3D cost volumes, and a cost volume network.
This framework naturally enables neural networks to utilize the information from corresponding pixels of reflection symmetry inside a single image.
We evaluate our method on the ShapeNet dataset \cite{chang2015shapenet} and real-world images. Extensive comparisons and analysis show that by detecting and utilizing symmetry structures, our method has better generalizability and accuracy, even when the object is not perfectly symmetric.

\section{Related Work}

\begin{figure}[t]
  \centering
  \subfloat[][2D reflection symmetry \label{fig:symmetry:2d}]{
    \includegraphics[width=0.30\linewidth]{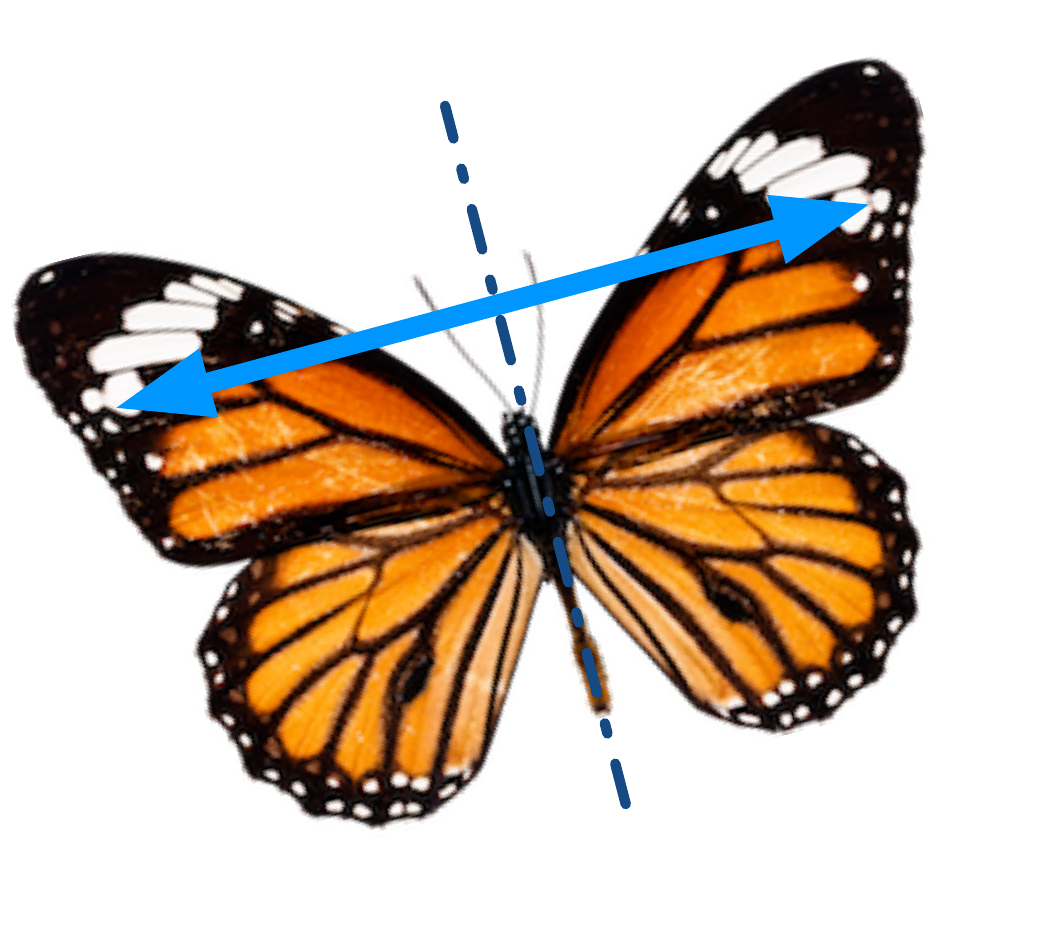}
  }
  \subfloat[][3D reflection symmetry \label{fig:symmetry:3d}]{
    \includegraphics[width=0.30\linewidth]{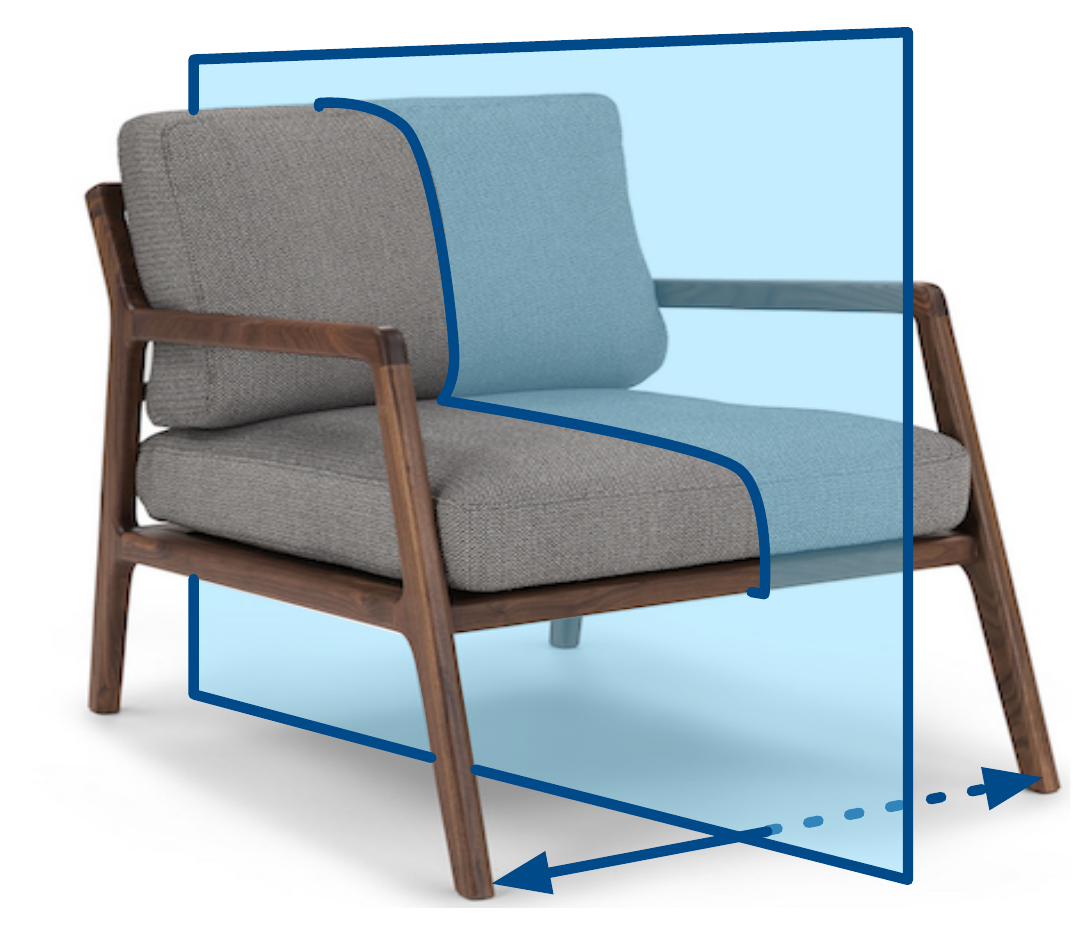}
  }
  \subfloat[][Scale ambiguity in 3D reconstruction \label{fig:symmetry:ambiguity}]{
    \includegraphics[width=0.38\linewidth]{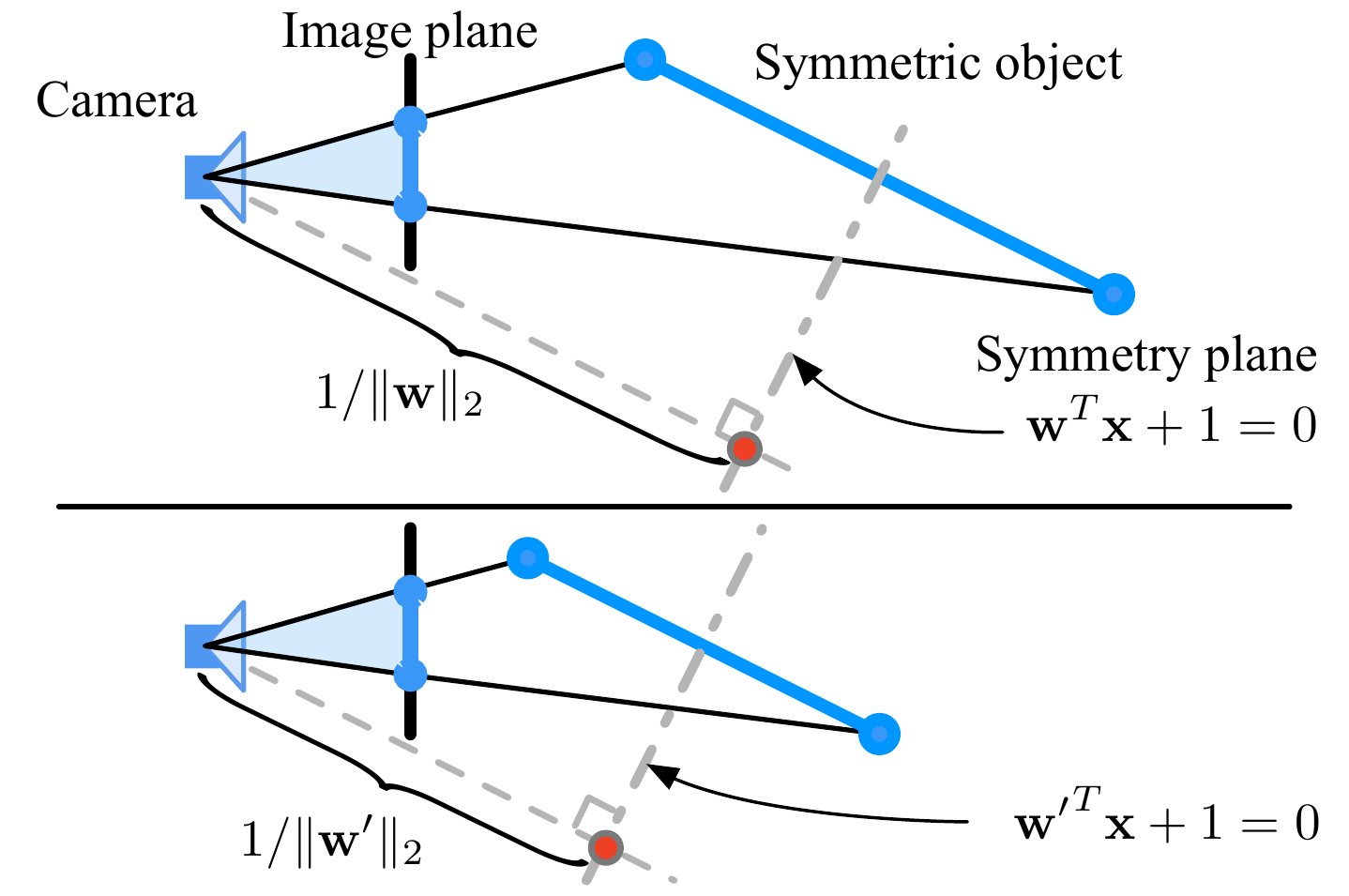}
  }
  \caption{Examples of 2D and 3D reflection symmetry and scale ambiguity in 3D reconstruction.  In (c), we show two scenes that only differ by a scale $c$.  The images of the two scenes are exactly the same, but the distances between the origin and two symmetry planes are different, i.e., $\|\w\|_2=c\|\w'\|_2$.}
  \label{fig:symmetry}
\end{figure}

\paragraph{Symmetry detection.}

For many years, scientists from vision science and psychology have found that symmetry plays an important role in human vision system \cite{Troje-symmetry,Vetter-symmetry2}.
Researchers in computer vision has exploited different kinds of symmetry for tasks such as recognition \cite{LiuY2010}, texture impainting \cite{korah2008analysis}, unsupervised shape recovering \cite{wu2019unsupervised}, and image manipulation \cite{zhang2020portrait}.
For 3D reconstruction, it is possible to reconstruct the 3D shape of an object from the correspondences of symmetry within \emph{a single image} (shape-from-symmetry) \cite{MundyJ1994,MaY2003,HongW2004,HongW2004-ECCV}.
However, detecting the symmetry and its correspondence points from images is challenging.
On one hand, most of geometry-based symmetry detection methods use handcrafted features and only work for 2D planar and front-facing objects \cite{loy2006detecting,JohanssonB2000-ICPR,MarolaG1989-PAMI,ZabrodskyH1995-PAMI,KiryatiN1998-IJCV}%
\footnote{A survey of related methods can be found in ICCV 2017 ``{\em Detecting Symmetry in the Wild}'' competition \href{https://sites.google.com/view/symcomp17}{https://sites.google.com/view/symcomp17}.}, as shown in \Cref{fig:symmetry:2d}.
The extracted 2D symmetry axes and correspondences cannot provide enough geometric cues for 3D reconstruction.
In order to make reflection symmetry useful for 3D reconstruction, it is necessary to detect the 3D mirror plane and corresponding points of symmetric objects (\Cref{fig:symmetry:3d}) from perspective images.
On the other hand, recent single-image camera pose detection neural networks \cite{chang2015shapenet,insafutdinov2018unsupervised,zhou2019continuity} can approximately recover the camera orientation with respect to the canonical pose, which gives mirror plane of symmetry.
However, the camera poses from those data-driven networks are not accurate enough, because they do not exploit the geometric constraints of symmetry.
To remedy the above issues, our SymmtreyNet takes the advantage of both worlds.
The proposed method first detects the 3D mirror plane of a symmetric object from an image and then recovers the depth map by finding the pixel-wise correspondence with respect to the symmetry plane, all of which are supported with geometric principles.
Our experiment (\Cref{sec:exp}) shows that SymmtreyNet is indeed more accurate for 3D symmetry plane detection, compared to previous learning-based methods \cite{zhou2019continuity,xu2019disn}.

\paragraph{Learning-based single-view 3D reconstruction.}
Inspired by the success of CNNs in classification and detection,
numerous 3D representations and associated learning schemes have been explored under the setting of single-view 3D reconstruction, including depth maps \cite{fu2018deep,chang2018pyramid}, voxels \cite{yan2016perspective,wu2018learning}, point clouds \cite{fan2017point}, signed distance fields (SDF) \cite{park2019deepsdf,mescheder2019occupancy}, and meshes \cite{groueix2018atlasnet,wang2018pixel2mesh}. 
Although these methods demonstrate promising results on some datasets, single-view reconstruction is essentially an {\em ill-posed} problem.
Without geometric constraints, inferred shapes {\em will not be accurate enough} by extrapolating from training data, especially for unseen objects.
To alleviate this issue, our method leverages the symmetry prior for accurate single-view 3D reconstruction.

\paragraph{Multi-view stereopsis.}
Traditional multi-view stereo methods build the cost volumes from photometric measures of images, regularize the cost volume, and post-process to recover the depth maps \cite{kang2001handling,furukawa2010towards,furukawa2015multi,schonberger2016pixelwise}.
Recent efforts leverage learning-based methods and have shown promising results on benchmarks \cite{geiger2013vision} with CNNs. 
Some directly learn patch correspondence between two~\cite{zbontar2016stereo} or more views~\cite{hartmann2017learned}. Others build plane-sweep cost volumes from image features and employ 3D CNNs to regularize cost volumes, which then can be either transformed into 3D representations~\cite{ji2017surfacenet,kar2017learning} or aggregated into depth maps~\cite{yao2018mvsnet,huang2018deepmvs,leroy2018shape,choi2018learning}.
Different from these methods, our approach builds the plane-sweep cost volume through a symmetry-based feature warping module, which make such powerful tool applicable to the single-image setting.

\section{Methods}

\subsection{Camera Model and 3D Symmetry} \label{sec:symmetry}

\begin{figure*}[t]
  \centering
  \includegraphics[width=0.99\linewidth]{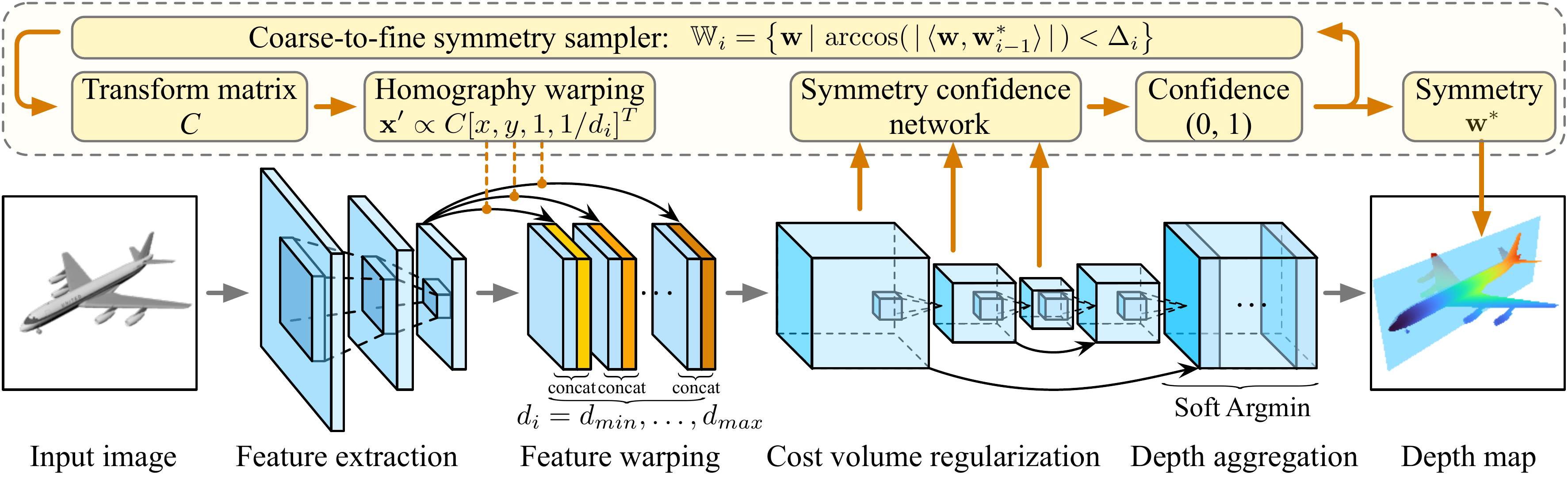}
  \caption{Overview of the SymmetryNet. Input images first go through the feature extraction network (backbone), down-sampled by a scale of 4. Features are warped by a warping module based on the symmetry transformation $\mathbf{C}$ and depth $d_i$. A cost volume is constructed by fusing the warped features and then feeding into a cost volume regularization network to compute the confidence of symmetry and the depth probability tensor. The final depth map is regressed by aggregating the probability tensor and upsampling to original resolution.
  }
  \label{fig:network}
\end{figure*}

Let $\object \subset \bbR^4$ be the set of points in the homogeneous coordinate that are on the surface of an object.  If we say $\object$ admits the \emph{symmetry}\footnote{An object might admit multiple symmetries. For example, a rectangle has two reflective symmetries and one rotational symmetry. We here describe one at a time.} with respect to a rigid transformation
$\M \in \bbR^{4 \times 4}$, it means that
\begin{equation}
    \forall \X \in \object: \M\X \in \object, \quad \mbox{and}\quad \feature(\X) = \feature(\M\X), \label{eq:sym}
\end{equation}
where $\X$ is homogeneous coordinates of a point on the surface of the object, $\M\X$ is the corresponding point of $\X$ with respect to the symmetry, and $\feature(\cdot)$ represents the surface properties at a given point, such as the surface material and texture.  For example, if an object has reflection symmetry with respect to the Y-Z plane in the world coordinate, then we have its transformation $\M_x = \mathrm{diag}(-1, 1, 1, 0)$.  \Cref{fig:symmetry} shows some examples of reflection symmetry.

Given two 3D points $\X,\X' \in \object$ in the homogeneous coordinate that are associated by the symmetry transform $\X' = \M \X$, their 2D projections $\x$ and $\x'$ must satisfy the following conditions:
\begin{equation}
  \x = \K\Rt\X/d,\quad \mbox{and} \quad  \x' = \K\Rt\X'/d'. \label{eq:proj2}
\end{equation}
Here, we keep all vectors in $\mathbb{R}^4$.  $\x=[x,y,1,1/d]^T$ and $\x'=[x',y',1,1/d']^T$ represent the 2D coordinates of the points in the pixel space, $d$ and $d'$ are the depth in the camera space, 
$\K \in \mathbb{R}^{4\times4}$ is the camera intrinsic matrix, and $
\Rt = \begin{bsmallmatrix}
  \R & \t \\
   0  & 1
\end{bsmallmatrix}
$ is the camera extrinsic matrix that rotates and translates the coordinate from the object space to the camera space.

From Equation \eqref{eq:proj2}, we can derive the following constraint for their 2D projections $\x$ and $\x'$:
\begin{equation}
  \x' \propto \underbrace{\K\Rt\M\Rt^{-1}\K^{-1}}_{\C} \x \doteq \C \x. \label{eq:correspondence}
\end{equation}
We use the proportional symbol here as the 3rd dimension of $\x'$ can always be normalized to one so the scale factor does not matter. The constraint in \Cref{eq:correspondence} is valuable to us because the neural network now has a geometrically meaningful way to check whether the estimated depth $d$ is reasonable at $(x,y)$ by comparing the image appearance at $(x, y)$ and $(x', y')$, where $(x', y')$ is computed from \Cref{eq:correspondence} given $x$, $y$, and $d$.  If $d$ is a good estimation, the two corresponding image patches should be similar due to $\feature(\X) = \feature(\X')$ from the symmetry constraint in \Cref{eq:sym}. This is often called \emph{photo-consistency} in the literature of multi-view steropsis \cite{furukawa2009accurate,tola2012efficient,yao2018mvsnet}.

An alternative way to understand \Cref{eq:correspondence} is to substitute $\X' = \M \X$ into Equation \eqref{eq:proj2} and treat the later equation as the projection from another view. By doing that, we reduce the problem of shape-from-symmetry to two-view steropsis, only that the stereo pair is in special positions \cite{MaY2003}.

\paragraph{Reflection symmetry.}
\Cref{eq:correspondence} gives us a generalized way to represent any types of symmetry with matrix $\C=\K\Rt\M\Rt^{-1}\K^{-1}$.  For reflection symmetry, a more intuitive parametrization is to use the equation of the symmetry plane in the camera space.  Let $\tilde \x \in \mathbb{R}^3$ be the coordinate of a point on the symmetry plane in the camera space.  The equation of the symmetry plane can be written as
\begin{equation}
\w^T \tilde{\x} + 1 = 0,
\end{equation}
where we use $\w\in\mathbb{R}^3$ as the parameterization of symmetry.  The relationship between $\C$ and $\w$ is
\begin{equation}
  \C = \K \left(\mathbf{I} - \frac{2}{\|\w\|_2^2}\begin{bmatrix}
  \w \\
  \mathbf{0} \\
  \end{bmatrix}
  \begin{bmatrix} \w^T& \mathbf{1} \end{bmatrix} \right)\K^{-1}. \label{eq:relationship}
\end{equation}
We derive \Cref{eq:relationship} in the supplementary material.  The goal of reflection symmetry detection is to recover $\w$ from images.

On the first impression, one may find it strange that $\Rt$ in \Cref{eq:correspondence} has 6 degrees of freedoms (DoFs) while $\w$ only has 3.  This is due to the specialty of reflection symmetry.  Rotating the camera with respect to the normal of the symmetry plane (1 DoF) and translating the camera along the symmetry plane (2 DoFs) will not change the location of the camera with respect to the symmetry plane.  Therefore the number of DoFs in reflection symmetry is indeed $6-1-2=3$.

\paragraph{Scale ambiguity.}
Similar to structure-from-motion in which it is impossible to determine the absolute size of scenes \cite{hartley2003multiple,MaY2003}, shape-from-symmetry also has a scale ambiguity.
This is demonstrated in \Cref{fig:symmetry:ambiguity}.
In the case of reflection symmetry, we cannot determine $\|\w\|_2$ (i.e., the symmetry plane's distance from origin) from correspondences in an image without relying on size priors.
This is because it is always possible to scale the scene by a constant (and thus scale $\|\w\|_2$) without affecting images.
Therefore, we choose not to recover $\|\w\|_2$ in SymmetryNet.
Instead, we fix $\|\w\|_2$ to be a constant and leave the ambiguity as it is.
For real-world applications, this scale ambiguity can be resolved when the object size or the distance between the object and the camera is known.

\subsection{Overall Pipeline of SymmetryNet} \label{sec:overall}

\Cref{fig:network} illustrates the overall neural network pipeline of SymmetryNet during inference. SymmetryNet fulfills two tasks simultaneously with an end-to-end deep learning framework: \emph{symmetry detection} and \textit{depth estimation}. It takes a single image as input and outputs parameters of the symmetry plane and depth maps.  We note that our method treats the symmetry as a prior rather than a hard constraint, so the pipeline still works for objects that are not perfectly symmetric.

\paragraph{Coarse-to-fine inference.} SymmetryNet finds the normal of the symmetry plane with a coarse-to-fine strategy.  In $i$th round of inference, the network first samples $N$ candidates symmetry plane $\{\w_{i}^k\}_{k=1}^{K}$ using the coarse-to-fine symmetry sampler (\Cref{sec:sampler}) and computes 2D feature maps  (\Cref{sec:backbone}).  For each symmetry pose, 2D feature maps are then used to assemble a 3D cost volume (\Cref{sec:warping}) for photo-consistency matching. After that, the confidence branch of the cost volume network (\Cref{sec:costvolume}) converts the cost volume tensor into a confidence value for each symmetry plane.  The pose with the highest confidence $\w_i^{*}$ is then fed back into the symmetry sampler and used to start the next round of inference.  After an accurate symmetry pose is pinpointed in the coarse-to-fine inference, the final depth map is computed by aggregating the depth probability map $\P$ from the depth branch of the cost volume network (\Cref{sec:costvolume}).

\begin{figure*}[t]
  \centering
  \subfloat[][Coarse-to-fine inference \label{fig:demo:c2f}]{%
  \begin{minipage}[b]{.4\textwidth}
  \setlength{\lineskip}{0pt}
  \frame{\includegraphics[width=0.49\linewidth]{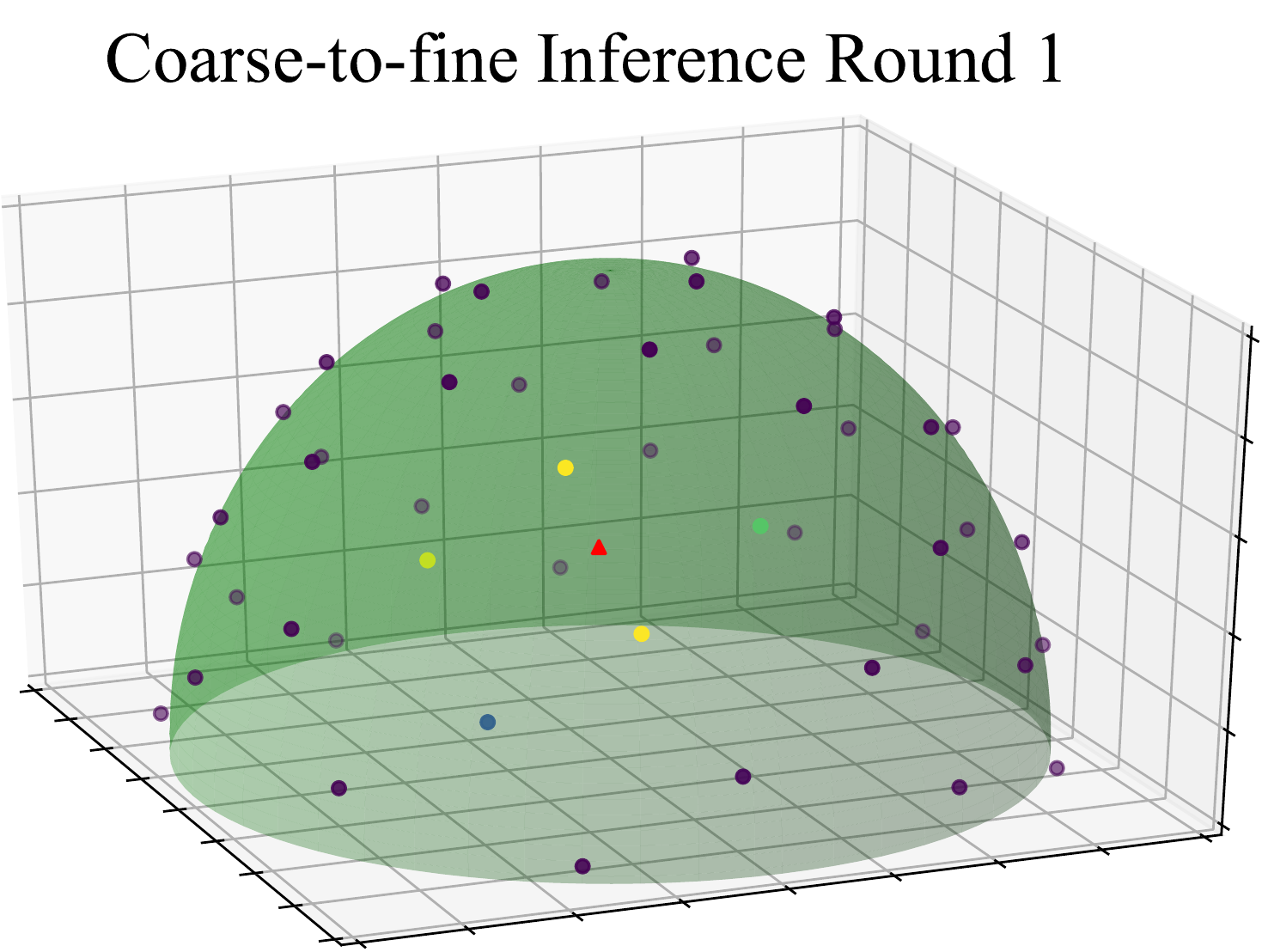}}%
  \frame{\includegraphics[width=0.49\linewidth]{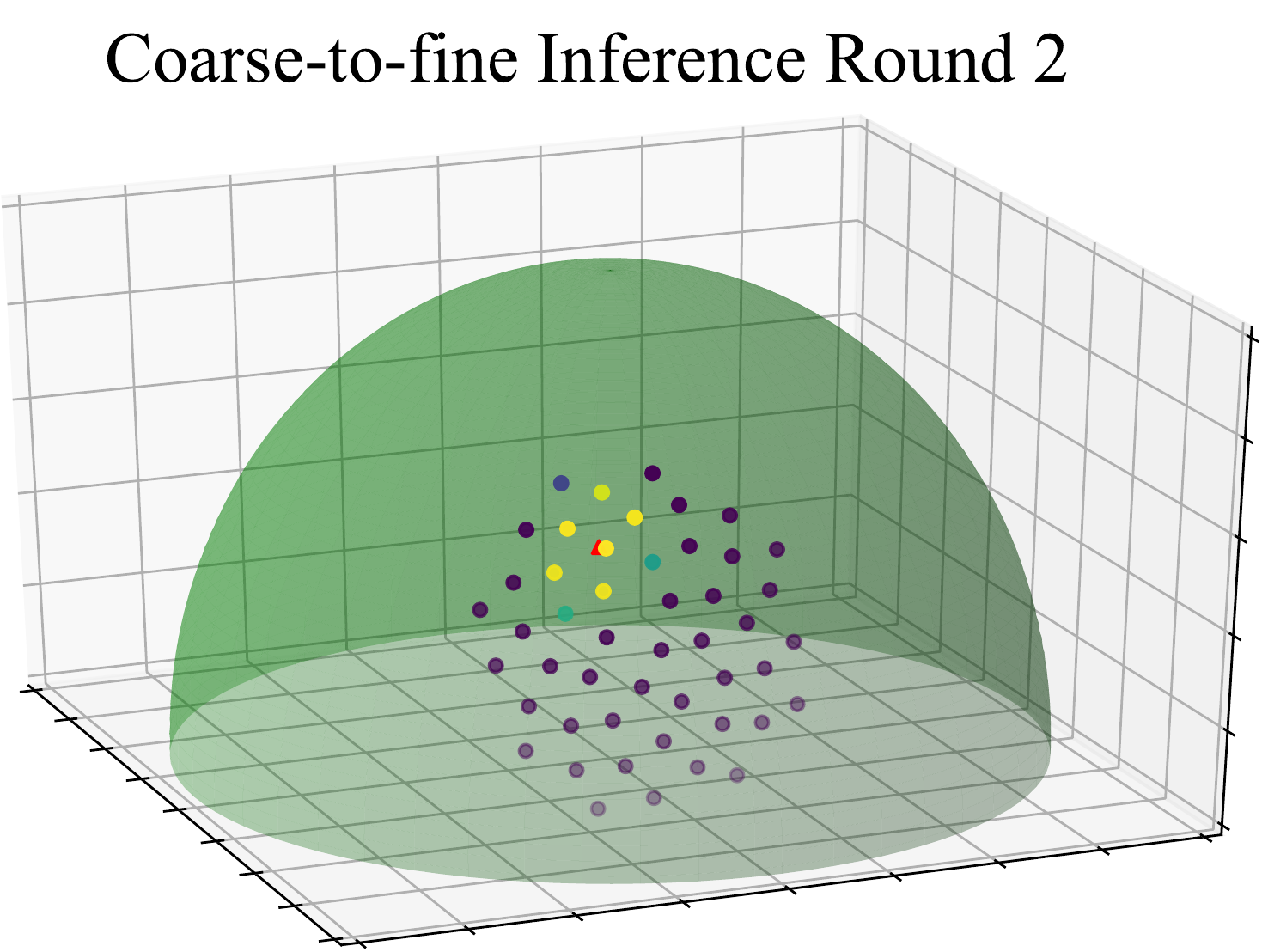}}
  \frame{\includegraphics[width=0.49\linewidth]{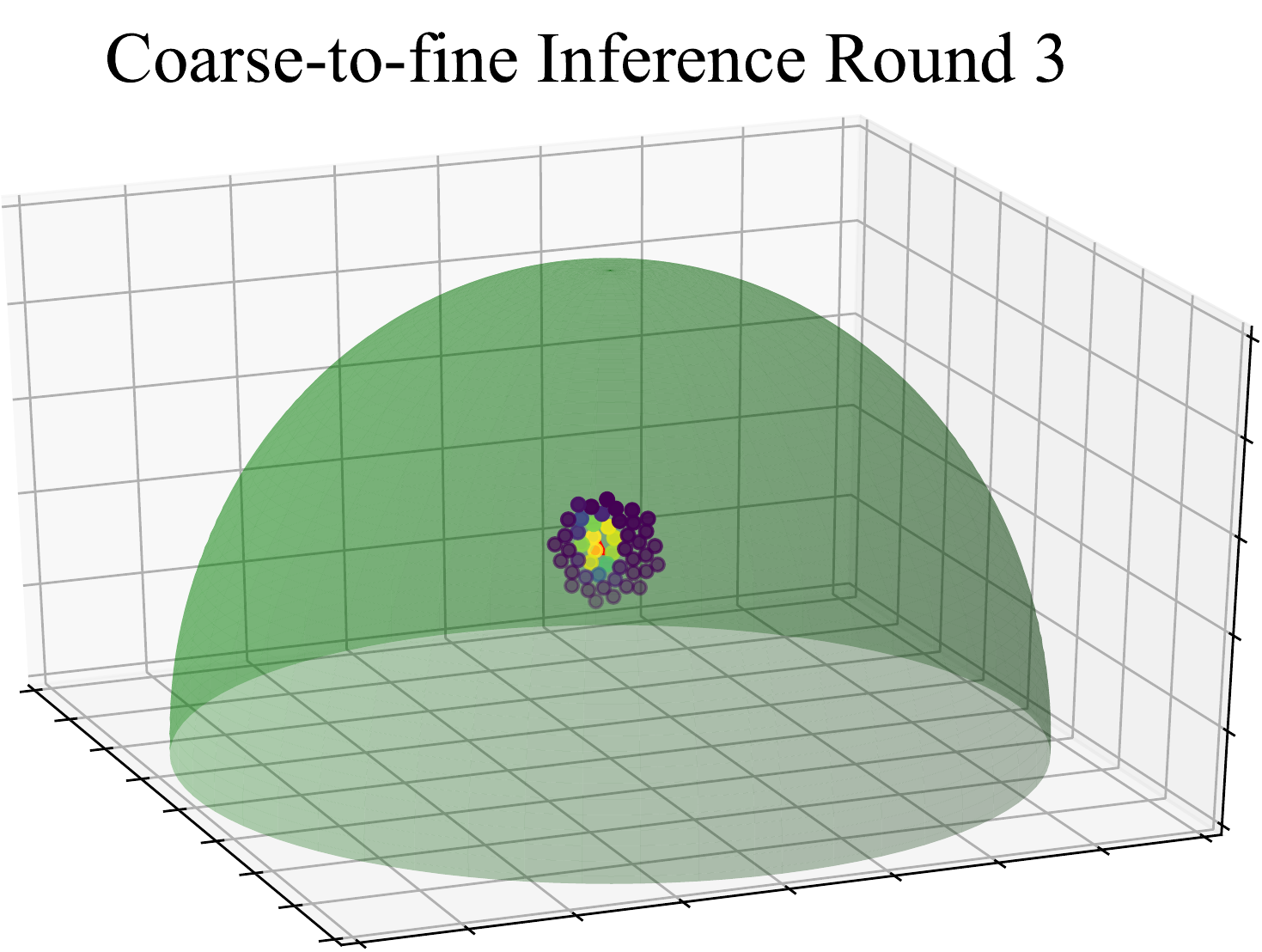}}%
  \frame{\includegraphics[width=0.49\linewidth]{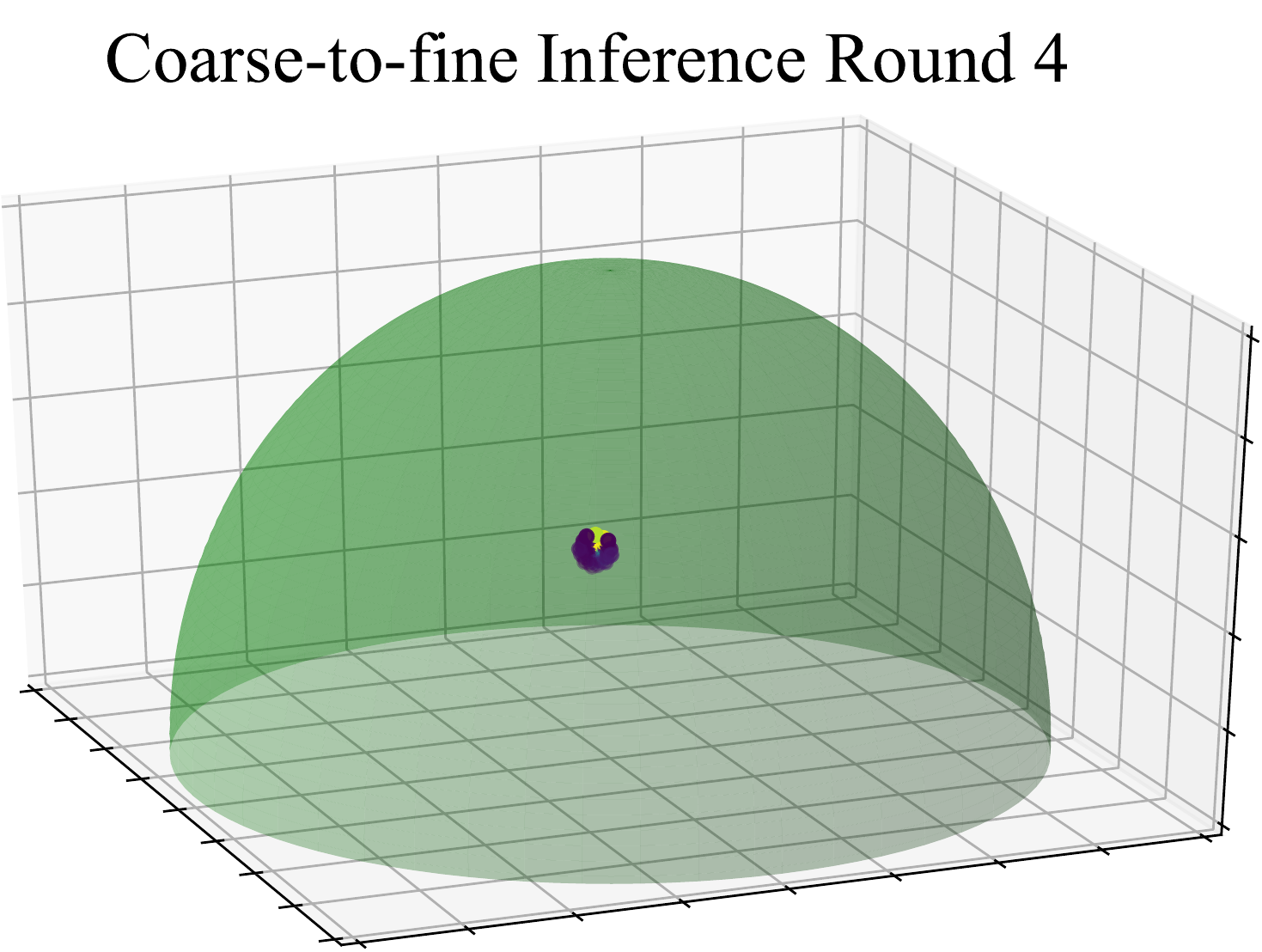}}%
  \end{minipage}%
  }
  \subfloat[][Feature warping module and cost volume \label{fig:demo:warping}]{
    \includegraphics[width=0.57\linewidth]{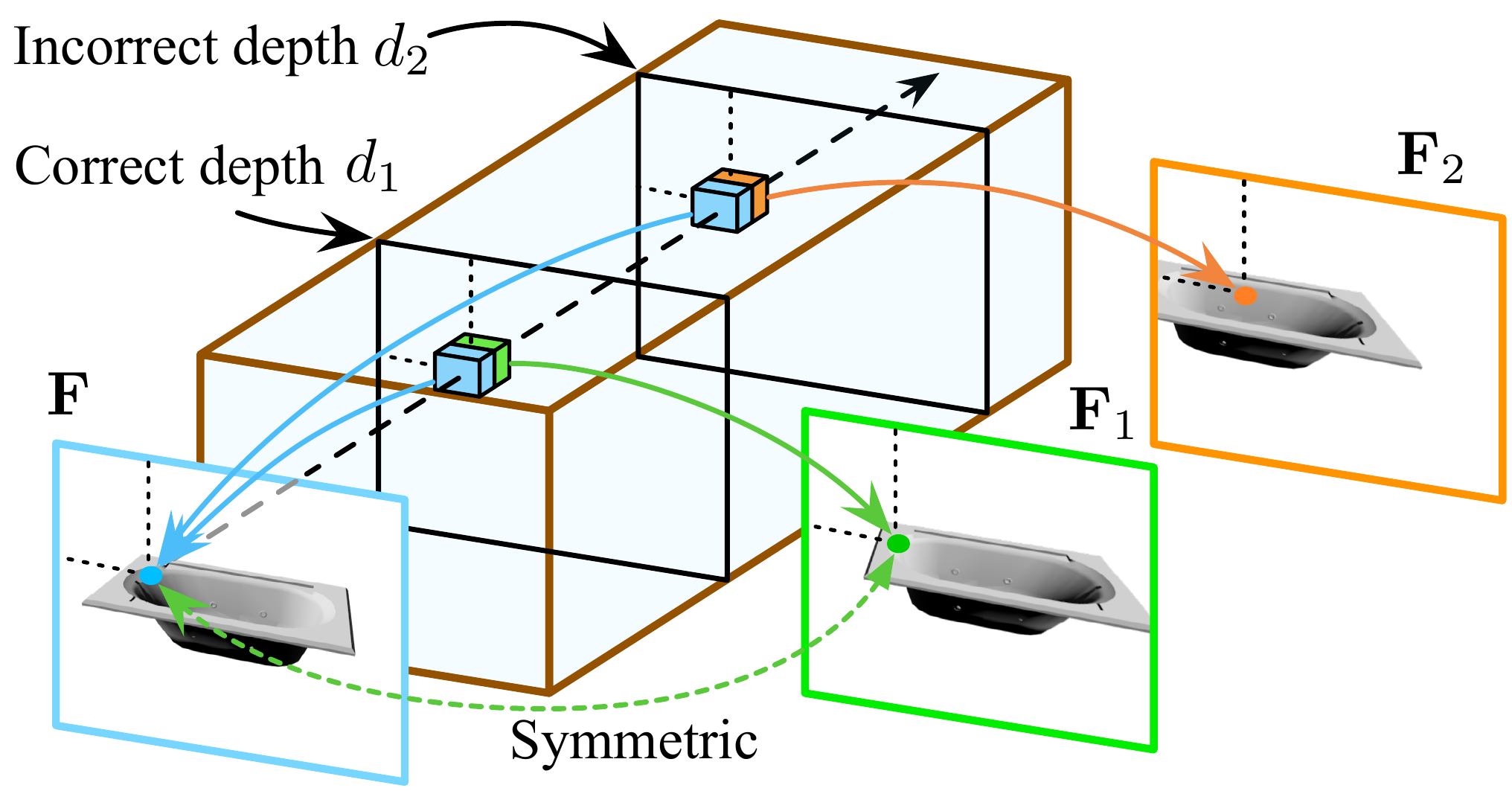}
  }
  \caption[]{Illustration of the process of coarse-to-fine inference of symmetry detection and feature warping module. (a): The sampled normal direction in a 4-round coarse-to-fine inference.  The color of points represents the scores from symmetry confidence network. (b): The feature map $\mathbf{F}$ is warped to $\mathbf{F}_i$ according to the transformation $\x' = \mathbf{C}\x$ for various depth $d_i$ in $\x$.  Here, the \textcolor{cyan}{input feature} (cyan dot) corresponds to warped features at \textcolor{Green}{correct depth} $d_1$ (green dot)  rather than the warped features at \textcolor{orange}{incorrect depth} $d_2$ (orange dot).}
  \label{fig:demo}
\end{figure*}

\subsection{Symmetry Sampler} \label{sec:sampler}
As shown in \Cref{fig:demo:c2f}, the symmetry sampler uniformly samples $\{\w_{i}^k\}_{k=1}^{K}$ from $\mathbb{W}_i \subset \mathbb{R}^3$, where $\mathbb{W}_i$ is the sampling space of the $i$th round of inference. For the first round, the candidates are sampled from the surface of the unit sphere, i.e., $\mathbb{W}_1 = \mathbb{S}^2$.  For the following rounds, we set $\mathbb{W}_i=\{\w\in\mathbb{S}^2\,|\,\arccos(|\langle\w,\w_{i-1}^*\rangle|) < \Delta_i\}$ to be a spherical cap, where $\w_{i-1}^*$ is the optimal $\w$ from the previous round and $\Delta_i$ is a hyper-parameter.  We use Fibonacci lattice \cite{gonzalez2010measurement} to generate a relatively uniform grid on the surface of a spherical cap, similar to the strategy used by \cite{zhou2019neurvps}.

\subsection{Backbone Network} \label{sec:backbone}
The goal of the {\em backbone network} is to extract 2D features from images.  We use a modified ResNet-like network as our backbone.  To reduce the memory footprint, we first down-sample the image with a stride-2 $5 \times 5$ convolution.  After that, the network has 8 \emph{basic blocks} \cite{he2016deep} with ReLU activation.  The 5th basic block uses stride-2 convolution to further downsample the feature maps.   The number of channels is 64.  The output feature map $\mathbf{F}$ has dimension $\lfloor \frac H 4\rfloor \times \lfloor \frac W 4\rfloor \times 64$.  The network structure diagram is shown in \Cref{fig:architecture} of the supplementary materials.

\subsection{Feature Warping Module} \label{sec:warping}

The function of the {\em feature warping module} is to construct the initial 3D cost volume tensor $\V(x,y,d)$ for photo-consistency matching.  We uniformly discretize $d$ so that $d \in \depthSet=\{d_{\min} + \frac{i}{D-1}(d_{\max}-d_{\min}) \mid i = 0,1,\dots,D-1\}$ to make the 3D cost volume homogeneous to 3D convolution, in which $d_{\min}$ and $d_{\max}$ is the minimal and maximal depth we want to predict and $D$ is the number of sampling points for depth. As mentioned in \Cref{sec:symmetry}, the correctness of $d$ at $(x, y)$ correlates with the appearance similarity of the image patch at pixels represented by $\x$ and $\C\x$.  Therefore, we set $\V$ by concatenating the backbone features at these two locations, i.e.,
\begin{equation}
    \V(x, y, d) = \Big[\mathbf{F}(x, y), \,\mathbf{F}(x', y')\Big], \quad \text{ where }\;
    [\,x', y', 1, 1/d'\,]^T  \propto \C [\,x, y, 1, 1/d\,]^T.
\end{equation}
Here $\mathbf{F}$ is the backbone feature, and $\C$ is computed from the sampled symmetry plane $\hat\w$.  We apply bilinear interpolation to access the features at non-integer coordinates.  The dimension of the cost volume tensor is $\lfloor \frac H 4\rfloor \times \lfloor \frac W 4\rfloor \times D \times 32$.

\subsection{Cost Volume Network} \label{sec:costvolume}
The cost volume network has two branches.
The first branch (confidence branch, labeled as \emph{symmetry confidence network} in \Cref{fig:network}) outputs the confidence of $\hat{\w}$ used by the feature warping module for coarse-to-fine inference.
The second branch (depth branch) outputs the estimated depth probability tensor $\P(x,y, d):=\Pr[\D(x,y)=d]$, where $\D$ is the ground truth depth map.
The cost volume aggregated from image features can be noisy, so we use a 3D hourglass-style network \cite{chang2018pyramid} that first down-samples the cost volume with 3D convolution (encoder), and then up-samples it back with transposed convolution (decoder).
For the confidence branch, we aggregate the multi-resolution encoder features with max-pool operators and then apply the sigmoid function to normalize the confidence values into $[0, 1]$.
For the depth probability branch, we apply a softmax operator along the depth dimension of the decoder features to compute the final depth probability tensor.

\paragraph{Depth aggregation.} \label{sec:aggregation}
We compute the expectation of depth from the probability tensor $\P$ as the depth map prediction $\hat\D$.  This is sometimes referred as \emph{soft argmin} \cite{kendall2017end}.  Mathematically, we have
\begin{align}
  \hat\D(x,y) &= \frac{1}{|\depthSet|} \sum_{d \in \depthSet} d\P(x,y, d). \label{eq:softargmin}
\end{align}

\vspace{-3mm}
\subsection{Training and Supervision} \label{sec:training}
During training, we sample symmetry planes for each image according to the hyper-parameter $\Delta_i$.
For the $i$th level,  symmetry poses are sampled from $\{\hat\w\in\mathbb{S}^2\,|\,\arccos(|\langle\w,\hat\w\rangle|) \le \Delta_i\}$,
where $\w$ is the ground truth symmetry pose.
We also add a random sample $\hat{\w} \in \mathbb{S}^2$ to reduce the sampling bias.
For each sampled $\hat\w$, its confidence labels is $l_{i}=\mathbf{1}[\arccos(|\langle\w,\hat\w\rangle|) < \Delta_{i}]$ for the $i$th level.

We use $\ell_1$ error as the supervision of depth and rescale the ground truth depth according to $\|\hat{\w}\|_2$. The training error could be written as $L = L_{\mathrm{dpt}} + L_{\mathrm{cls}}$, where
\begin{equation}
    L_{\mathrm{dpt}} = \frac{1}{n}\sum_{x,y} \left|\hat\D(x,y) - \frac{\|\hat{\w}\|_2}{\|\w\|_2}\D(x,y)\right| \quad  \text{ and } \quad L_{\mathrm{cls}} = \sum_{i}\mathrm{BCE}(\hat l_{i}, l_{i}).
\end{equation}
Here $n$ is the number of pixels, $\w$ is the ground truth plane parameter, $\mathrm{BCE}$ represents the binary cross entropy error, and $\hat l_i$ is the confidence of $\hat{w}$ for the $i$th level of coarse-to-fine inference.  %

\section{Experiments} \label{sec:exp}

\subsection{Settings}
\paragraph{Datasets.}
We conduct experiments on the ShapeNet dataset \cite{chang2015shapenet}, in which models have already been processed so that in their canonical poses the Y-Z plane is the plane of the reflection symmetry.
We use the same camera pose, intrinsic, and train/validation/test split from a 13-category subset of the dataset as in \cite{kar2017learning,wang2018pixel2mesh,choy20163d} to make the comparison easy and fair.
We do \emph{not} filter out any asymmetric objects.
We use Blender to render the images with resolution $256 \times 256$.

\paragraph{Implementation details.}
We implement SymmetryNet in PyTorch.
We use the plane $x=0$ in the world space as the ground truth symmetry plane because it is explicitly aligned for each model by authors of ShapeNet.
We set $d_{\min}=0.64$ and $d_{\max}=1.23$ according to the depth distribution of the dataset.
We set $D=64$ for the depth of the cost volume.
We use $K=4$ rounds in the coarse-to-fine inference.
We set $\Delta=[20.7^\circ, 6.44^\circ, 1.99^\circ, 0.61^\circ]$
Our experiments are conducted on three NVIDIA RTX 2080Ti GPUs.  We use the Adam optimizer \cite{kingma2014adam} for training.
Learning rate is set to $3 \times 10^{-4}$ and batch size is set to 16 per GPU.
We train the SymmetryNet for 6 epochs and decay the learning rate by a factor of 10 at 4th epoch.
The overall inference speed is about 1 image per second on a single GPU.

\paragraph{Symmetry detection.}  We compare the accuracy of our symmetry plane procedure with the state-of-the-art single-view camera pose detection methods \cite{zhou2019continuity}.
Because the object in the world space is symmetric with respect the Y-Z plane, we can convert the results of single-view camera pose detection methods into $\w$ of the symmetry plane in the camera space.
Our baseline approach is the recent \cite{zhou2019continuity}, which identifies a 6D rotation representation that are more suitable for learning and uses neural network to regress that representation directly.
DISN~\cite{xu2019disn} implements \cite{zhou2019continuity} for ShapeNet. We report the performance of the pre-trained model from DISN on their rendering as our baseline.

For each image, we calculate the angle error between the ground truth $\w$ and estimated $\hat\w$ and plot the \emph{error-percentage} curve for each method, in which the x-axis represents the magnitude of the error (in degrees) and the y-axis shows the percentage of testing data with smaller error.

\paragraph{Depth estimation.}  We introduce Pixel2Mesh~\cite{wang2018pixel2mesh} and DISN~\cite{xu2019disn} as the baseline for depth estimation.
We use the pre-trained model provided by the authors and test on their own renderings that have the same camera setting as ours.
The resulting meshes are then rendered to depth maps using Blender.
Because meshes from Pixel2Mesh and DISN may not necessarily cover all the valid pixels, we only compute the error statistics for the pixels that are in the intersection of the silhouettes of predicted meshes and the ground truth image masks.

Besides regular testing, we introduce additional experiments to evaluate the generalizability of our approaches.  We train SymmetryNet only on three categories: cars, planes, and chairs, and test it on the rest 10 categories. To the best of our knowledge, there is no public benchmark with similar setting on the rendering of \cite{choy20163d}.  So we train a hourglass depth estimation network \cite{newell2016stacked} as our baseline.  We set its learning rate and number of epochs to be the same as those of SymmetryNet.

To evaluate the quality of depth maps, we report SILog, absRel, sqRel, and RMSE errors as in the KITTI depth prediction competition \cite{Uhrig2017THREEDV}.  To better understand the error distribution, we also plot the error-percentage curves for the average $\ell_1$ error and SILog~\cite{eigen2014depth}, similar to the ones described in the ``symmetry detection'' paragraph.  Because shape-from-symmetry has a scale ambiguity (\Cref{sec:symmetry}),  scale-invariant SILog \cite{eigen2014depth} is a more natural metric.  For other metrics that require absolute magnitude of depth, we re-scale the depth map using the ground truth $\|\w\|_2$ for error computing.

\begin{figure}[t]
    \newcommand{\lengthb}{0.33\linewidth}
    \centering

    \subfloat[][Plane normal error (in degrees)\label{fig:benchmark:pose}]{
      \includegraphics[width=\lengthb]{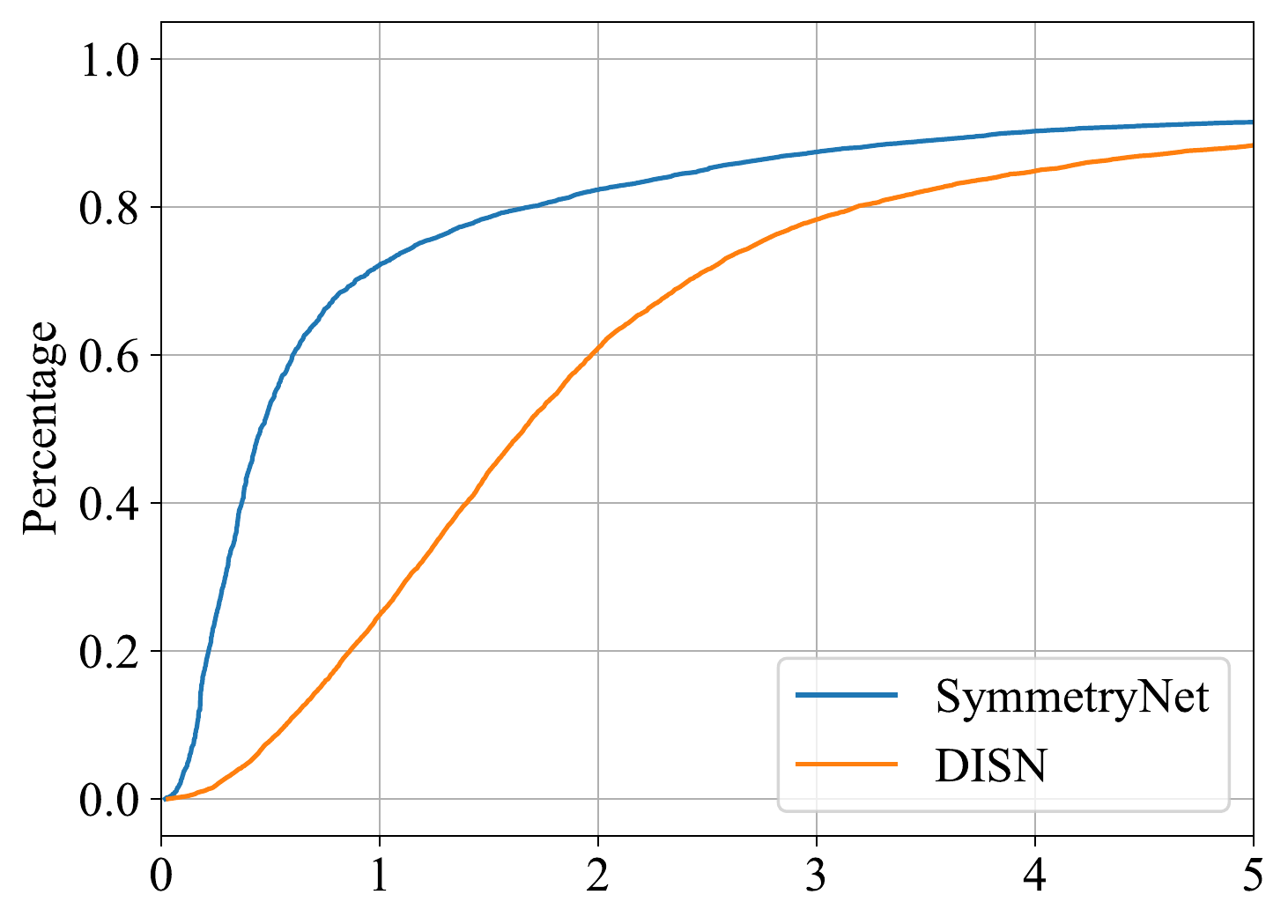}
    }%
    \subfloat[][Depth error (average $\ell_1$)\label{fig:benchmark:l1}]{
      \includegraphics[width=\lengthb]{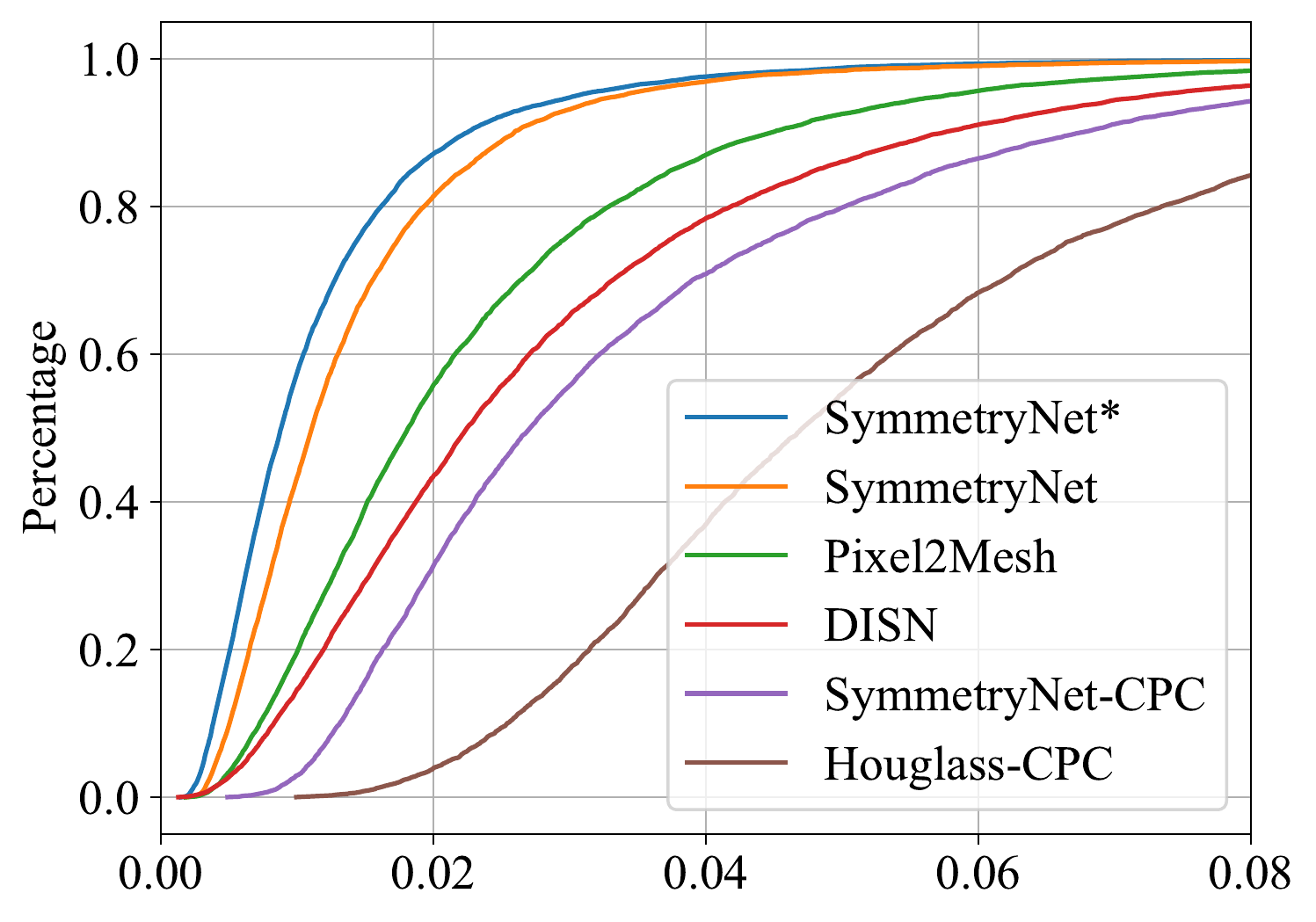}
    }%
    \subfloat[][Depth error (SILog~\cite{eigen2014depth})\label{fig:benchmark:sil}]{
      \includegraphics[width=\lengthb]{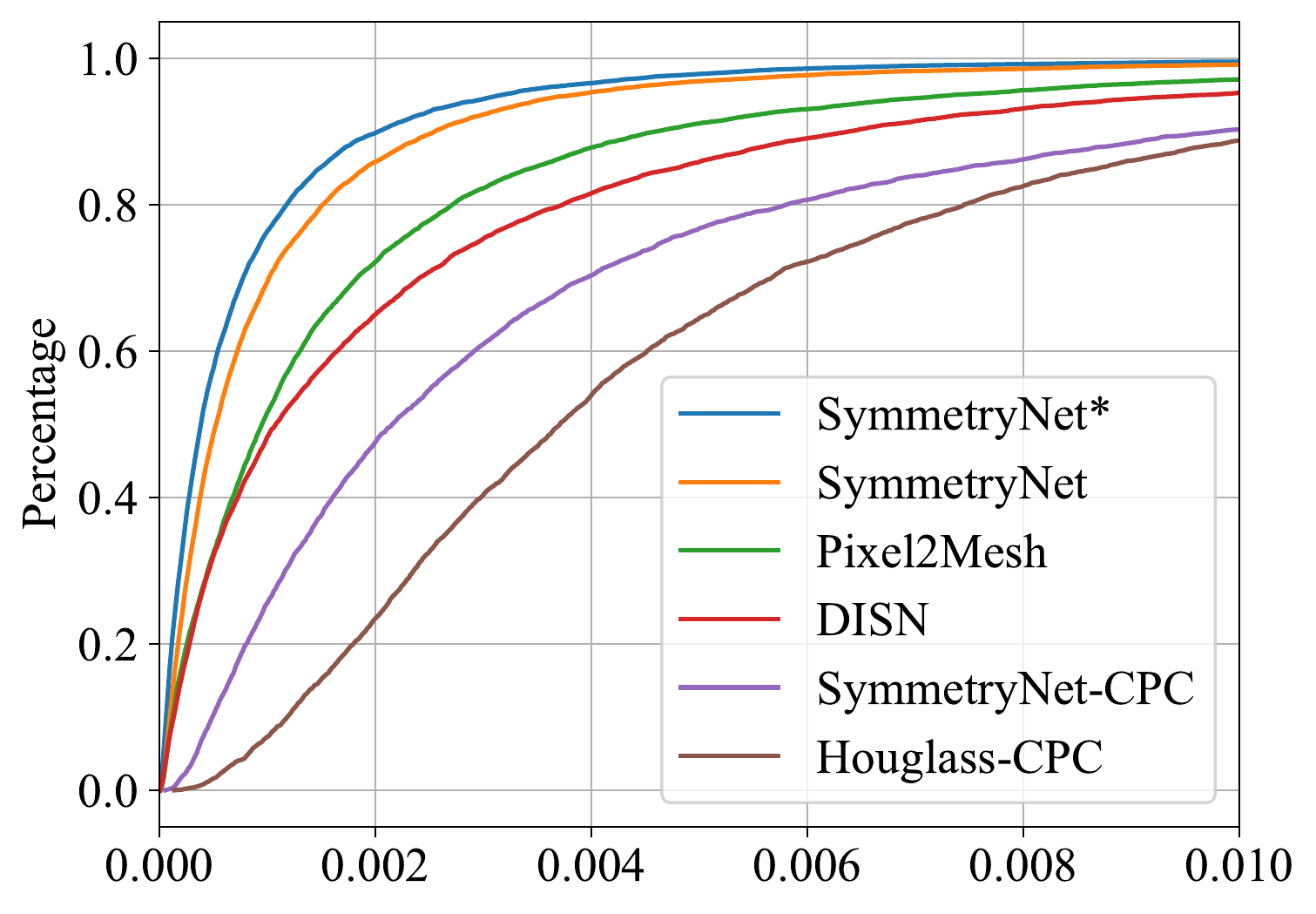}
    }
    \caption{Error-percentage curves of DISN \cite{xu2019disn}, Pixel2Mesh \cite{wang2018pixel2mesh}, and our SymmetryNet on ShapeNet.  Results with ``-CPC'' suffices are trained on images of cars, planes, and chairs, but tested on the other categories.  SymmetryNet* uses the ground truth symmetry plane as input. \emph{Higher curves are better.}}
    \label{fig:benchmark}
\end{figure}

\subsection{Results}
The results on the task of \emph{symmetry detection} are shown in \Cref{fig:benchmark:pose}.
By utilizing geometric cues from symmetry, our approach significantly outperforms the state-of-the-art camera pose detection network from \cite{xu2019disn} that directly regresses the camera pose from images.
The performance gap is even larger in the region of higher precision.
For example, SymmetryNet can achieve the accuracy within $1^\circ$ of error on about $72\%$ of testing data, while \cite{xu2019disn} can only achieve that kind of accuracy on about $24\%$ of testing data.
Such phenomena indicates that our symmetry-aware network is able to recover the normal of symmetry plane more accurately, while naive CNNs can only roughly determine the plane normal by interpolating from training data.

\begin{wraptable}{r}{.5\linewidth}

\vspace{-12pt}
\setlength{\tabcolsep}{1.5mm}
\renewcommand{\arraystretch}{1.2}
\resizebox{\linewidth}{!}{%
\begin{tabular}{|c|cccc|}
\hline
                                       & SILog           & absRel         & sqRel           & RMSE           \\ \hline
DISN \cite{xu2019disn}                 & 0.0025          & 0.040          & 0.0030          & 0.038          \\
Pixel2Mesh \cite{wang2018pixel2mesh}   & 0.0019          & 0.032          & 0.0022          & 0.032          \\
\textbf{SymmetryNet}                   & \textbf{0.0011} & \textbf{0.019} & \textbf{0.0009} & \textbf{0.021} \\ \hline
\textbf{SymmetryNet*}                  & \textbf{0.0008} & \textbf{0.016} & \textbf{0.0007} & \textbf{0.018} \\ \hline
Hourglass-CPC \cite{newell2016stacked} & 0.0051          & 0.080          & 0.0081          & 0.066          \\
\textbf{SymmetryNet-CPC}               & \textbf{0.0040} & \textbf{0.049} & \textbf{0.0038} & \textbf{0.044} \\ \hline
\end{tabular}%
}
\caption{Depth error of SymmetryNet and other baseline methods on ShapeNet. Results with ``-CPC'' suffices are trained on images of cars, planes, and chairs, but tested on the images from other categories.  SymmetryNet* uses the ground truth symmetry plane as input. \emph{Lower is better}.
}
\label{tab:comparison}
\vspace{2mm}

\end{wraptable}

The results on the task of \emph{depth estimation} are shown in \Cref{fig:benchmark:l1,fig:benchmark:sil} and \Cref{tab:comparison}.
On the regular setting, SymmetryNet outperforms DISN and Pixel2Mesh in all of the tested metrics.
In addition, SymmetryNet*, the variant of SymmetryNet that uses the ground truth symmetry plane instead of the one predicted in coarse-to-fine inference, only slightly outperforms the standard SymmetryNet.
These behaviors indicate that detecting symmetry planes and incorporating photo-consistency priors of reflection symmetry into the neural network makes the task of single-view reconstruction less ill-posed and thus can improve the performance.
We also note that DISN and Pixel2Mesh are \emph{strong baselines} because they are able to beat other popular methods on ShapeNet, such as R2N2~\cite{choy20163d}, AtlasNet~\cite{groueix2018atlasnet}, and OccNet~\cite{mescheder2019occupancy}.

In the setting of testing on images of unseen categories (``-CPC''), SymmetryNet is able to surpass our end-to-end depth regression baseline \cite{newell2016stacked}. We think this is because the introduced prior knowledge of reflection symmetry is able to help neural networks to generalize better on novel objects that are not seen in the training set.

\begin{figure*}[t]
    \centering
    \includegraphics[width=\linewidth]{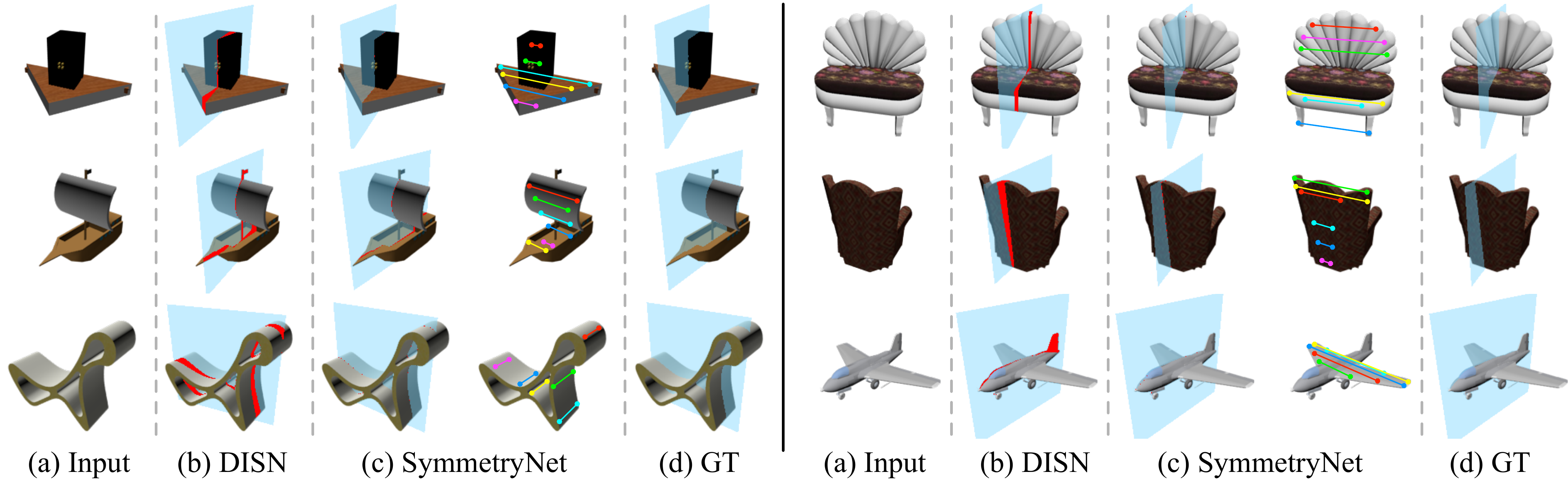}
    \caption{Qualitative results on the task of symmetry detection.  We show the detected symmetry planes from DISN \cite{xu2019disn} and our SymmetryNet.  Errors of symmetry planes, i.e., pixels that are between the predicted plane and the ground truth plane, are highlighted in red.  For SymmetryNet, we also visualize the sampled corresponding pixels from the depth probability tensor $\P$.}
    \label{fig:vis-symmetry}
\end{figure*}

\paragraph{Visualization.}
We visualize the detected symmetry in \Cref{fig:vis-symmetry}. 
We have the following observations:
1) our method outperforms previous methods on unusual objects, e.g. chairs in atypical shapes.
This indicates that learning-based methods need to extrapolate from seen patterns and cannot generalize to unusual images well, while our method relies more on geometry cues from symmetry, a more reliable source of information for 3D understanding.
2) Our method gives accurate symmetry planes even on challenging camera poses such as the orientation from the back of chairs.
We believe that this is because although visual cues along may not be remarkable enough in these cases, geometric information from correspondence helps to pinpoint the normal of symmetry planes.

\begin{figure*}[t]
    \centering
    \includegraphics[width=\linewidth]{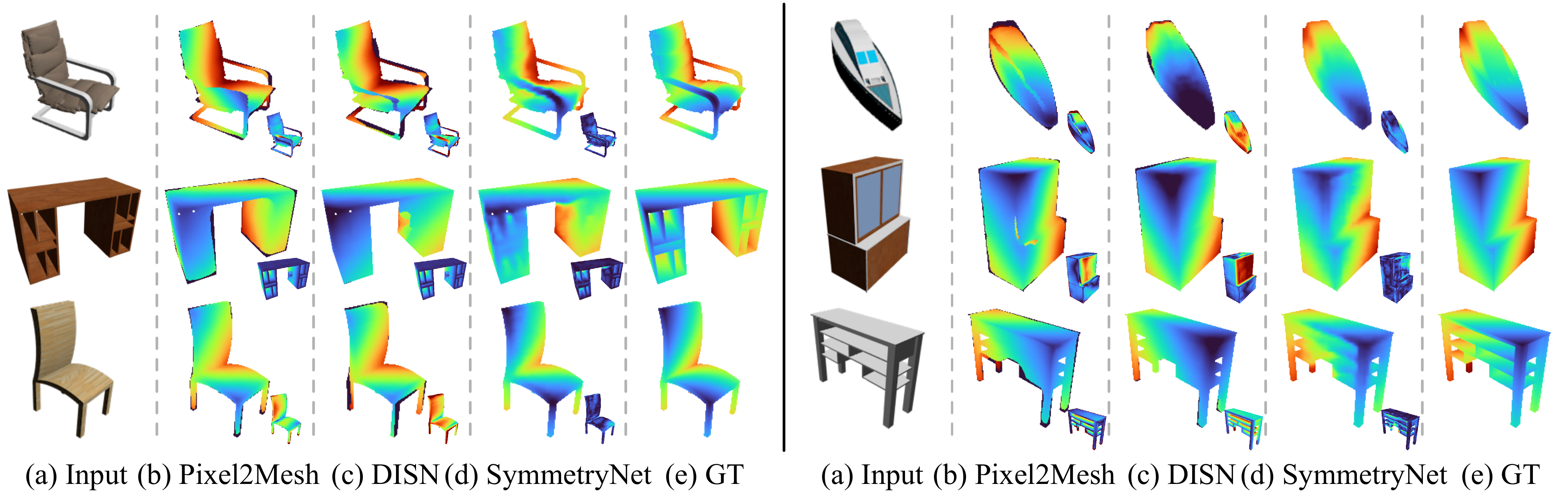}
    \caption{Qualitative results on the task of of depth estimation.  We visualize the depth maps from Pixel2Mesh \cite{wang2018pixel2mesh}, DISN \cite{xu2019disn}, and our SymmetryNet on ShapeNet.  The per-pixel $\ell_1$ errors are plotted at the lower right corner of each depth map.  Bluish color represents smaller values for error and depth.}
    \label{fig:visualization}
\end{figure*}

In \Cref{fig:visualization}, we show sampled depth maps of SymmetryNet and other baseline methods.
Visually, SymmetryNet gives the sharpest results among all the tested methods.
For example, it is able to capture the details of desk frames and the shapes of ship cabins, while those details are more blurry in the results of other methods.
In addition, in the region such as the chair armrests and table legs, SymmetryNet can recover the depth much more accurate compared to the baseline methods.
This is because for SymmetryNet, pixel-matching  based on photo-consistency in those areas is easy and can provide a strong signal, while other baseline methods need to extrapolate from the training data.  

\begin{figure}[b]
\vspace{-2mm}
    \centering
    \includegraphics[width=\linewidth]{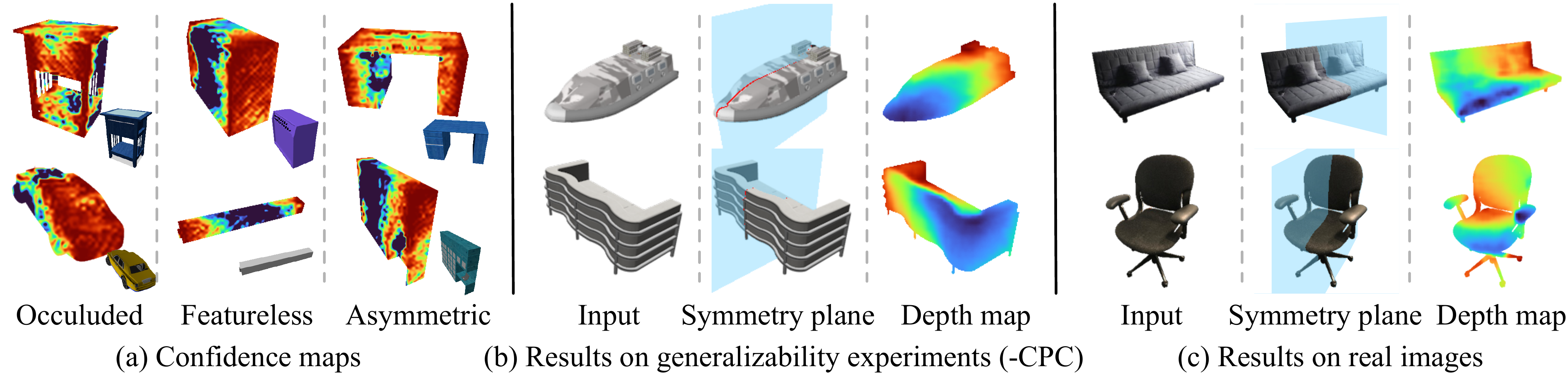}
    \caption{Additional visualization of SymmetryNet. (a) Depth confidence maps of SymmetryNet. Reddish color represents high confidence.  (b) Qualitative results of SymmetryNet on the images of unseen categories on ShapeNet.  (c) Qualitative results of ShapeNet-trained SymmetryNet tested on segmented real-world images.}
    \label{fig:confidence}
\end{figure}

To see how SymmetryNet generalizes on the unseen data, we visualize its results in the generalizability experiments in \Cref{fig:confidence}b. Although the training set does not contain any ships and sofas, SymmetryNet can still predict the good-quality symmetry planes and depth maps.  Finally, we test SymmetryNet on images from the real world.  As seen in  \Cref{fig:confidence}c, SymmetryNet trained on synthetic data can also generalize to some real-world images.


\paragraph{Depth Confidence Map.}
To further verify that SymmetryNet helps depth estimation with geometric cues from symmetry, we plot the depth confidence map \cite{yao2018mvsnet} by averaging four nearby depth hypotheses for each pixel in $\P$ in \Cref{fig:confidence}a.  We find three common cases when SymmetryNet is uncertain about its prediction: (1) the corresponding part of some pixels is occluded; (2) the object is asymmetric; (3) pixels are in textureless regions. In the first two cases, it is impossible for the neural network to find corresponding points when the photo-consistency assumption is violated.  The third case often occurs in the middle of objects. The image appearance towards the reflection plane is becoming increasingly harder for the neural network to establish correct correspondence, due to the lack of distinguishable shapes or textures. All the above cases match our expectation, which  supports our hypothesis that SymmetryNet indeed learns to use geometric cues.

\section{Discussion and Future Work}
In this work, we present a geometry-based learning framework to detect the reflection symmetry for 3D reconstruction.
However, our method is not just limited to the reflection symmetry.
With proper datasets and annotations, we can easily change $\M$ in \Cref{eq:correspondence} to adapt our method to other types of symmetry such as rotational symmetry and translational symmetry.
Currently, our method only supports depths maps as the output.
In the future, we will explore to predict more advanced shape representation, including voxels, meshes, and SDFs.

\bibliographystyle{plain}
\bibliography{paper}

\begin{thebibliography}{10}

\bibitem{chang2015shapenet}
Angel~X Chang, Thomas Funkhouser, Leonidas Guibas, Pat Hanrahan, Qixing Huang,
  Zimo Li, Silvio Savarese, Manolis Savva, Shuran Song, Hao Su, et~al.
\newblock {ShapeNet}: An information-rich {3D} model repository.
\newblock {\em arXiv preprint arXiv:1512.03012}, 2015.

\bibitem{chang2018pyramid}
Jia-Ren Chang and Yong-Sheng Chen.
\newblock Pyramid stereo matching network.
\newblock In {\em Proceedings of the IEEE Conference on Computer Vision and
  Pattern Recognition}, pages 5410--5418, 2018.

\bibitem{MultiView-chen2019point}
Rui Chen, Songfang Han, Jing Xu, and Hao Su.
\newblock Point-based multi-view stereo network.
\newblock In {\em Proceedings of the IEEE International Conference on Computer
  Vision}, pages 1538--1547, 2019.

\bibitem{choi2018learning}
Sungil Choi, Seungryong Kim, Kihong Park, and Kwanghoon Sohn.
\newblock Learning descriptor, confidence, and depth estimation in multi-view
  stereo.
\newblock In {\em Proceedings of the IEEE Conference on Computer Vision and
  Pattern Recognition Workshops}, pages 276--282, 2018.

\bibitem{choy20163d}
Christopher~B Choy, Danfei Xu, JunYoung Gwak, Kevin Chen, and Silvio Savarese.
\newblock {3D-R2N2}: A unified approach for single and multi-view 3d object
  reconstruction.
\newblock In {\em European conference on computer vision}, pages 628--644.
  Springer, 2016.

\bibitem{eigen2014depth}
David Eigen, Christian Puhrsch, and Rob Fergus.
\newblock Depth map prediction from a single image using a multi-scale deep
  network.
\newblock In {\em Advances in neural information processing systems}, pages
  2366--2374, 2014.

\bibitem{fan2017point}
Haoqiang Fan, Hao Su, and Leonidas~J Guibas.
\newblock A point set generation network for {3D} object reconstruction from a
  single image.
\newblock In {\em Proceedings of the IEEE conference on computer vision and
  pattern recognition}, pages 605--613, 2017.

\bibitem{fu2018deep}
Huan Fu, Mingming Gong, Chaohui Wang, Kayhan Batmanghelich, and Dacheng Tao.
\newblock Deep ordinal regression network for monocular depth estimation.
\newblock In {\em Proceedings of the IEEE Conference on Computer Vision and
  Pattern Recognition}, pages 2002--2011, 2018.

\bibitem{furukawa2010towards}
Yasutaka Furukawa, Brian Curless, Steven~M Seitz, and Richard Szeliski.
\newblock Towards internet-scale multi-view stereo.
\newblock In {\em 2010 IEEE computer society conference on computer vision and
  pattern recognition}, pages 1434--1441. IEEE, 2010.

\bibitem{furukawa2015multi}
Yasutaka Furukawa, Carlos Hern{\'a}ndez, et~al.
\newblock Multi-view stereo: A tutorial.
\newblock {\em Foundations and Trends{\textregistered} in Computer Graphics and
  Vision}, 9(1-2):1--148, 2015.

\bibitem{furukawa2009accurate}
Yasutaka Furukawa and Jean Ponce.
\newblock Accurate, dense, and robust multiview stereopsis.
\newblock {\em IEEE transactions on pattern analysis and machine intelligence},
  32(8):1362--1376, 2009.

\bibitem{MultiView-galliani2016just}
Silvano Galliani and Konrad Schindler.
\newblock Just look at the image: viewpoint-specific surface normal prediction
  for improved multi-view reconstruction.
\newblock In {\em Proceedings of the IEEE Conference on Computer Vision and
  Pattern Recognition}, pages 5479--5487, 2016.

\bibitem{geiger2013vision}
Andreas Geiger, Philip Lenz, Christoph Stiller, and Raquel Urtasun.
\newblock Vision meets robotics: The {KITTI} dataset.
\newblock {\em The International Journal of Robotics Research},
  32(11):1231--1237, 2013.

\bibitem{gonzalez2010measurement}
{\'A}lvaro Gonz{\'a}lez.
\newblock Measurement of areas on a sphere using fibonacci and
  latitude--longitude lattices.
\newblock {\em Mathematical Geosciences}, 2010.

\bibitem{groueix2018atlasnet}
Thibault Groueix, Matthew Fisher, Vladimir~G Kim, Bryan~C Russell, and Mathieu
  Aubry.
\newblock {AtlasNet}: A papier-mache approach to learning {3D} surface
  generation.
\newblock {\em arXiv preprint arXiv:1802.05384}, 2018.

\bibitem{hartley2003multiple}
Richard Hartley and Andrew Zisserman.
\newblock {\em Multiple view geometry in computer vision}.
\newblock Cambridge university press, 2003.

\bibitem{hartmann2017learned}
Wilfried Hartmann, Silvano Galliani, Michal Havlena, Luc Van~Gool, and Konrad
  Schindler.
\newblock Learned multi-patch similarity.
\newblock In {\em Proceedings of the IEEE International Conference on Computer
  Vision}, pages 1586--1594, 2017.

\bibitem{he2016deep}
Kaiming He, Xiangyu Zhang, Shaoqing Ren, and Jian Sun.
\newblock Deep residual learning for image recognition.
\newblock In {\em Proceedings of the IEEE conference on computer vision and
  pattern recognition}, pages 770--778, 2016.

\bibitem{HongW2004}
W.~Hong, A.~Y. Yang, and Y.~Ma.
\newblock On group symmetry in multiple view geometry: structure, pose and
  calibration from single images.
\newblock {\em International Journal of Computer Vision (IJCV)}, 60(3), 2004.

\bibitem{HongW2004-ECCV}
W.~Hong, Y.~Yu, and Y.~Ma.
\newblock Reconstruction of {3D} symmetric curves from perspective images
  without discrete features.
\newblock In {\em ECCV}, 2004.

\bibitem{huang2018deepmvs}
Po-Han Huang, Kevin Matzen, Johannes Kopf, Narendra Ahuja, and Jia-Bin Huang.
\newblock {DeepMVS}: Learning multi-view stereopsis.
\newblock In {\em Proceedings of the IEEE Conference on Computer Vision and
  Pattern Recognition}, pages 2821--2830, 2018.

\bibitem{insafutdinov2018unsupervised}
Eldar Insafutdinov and Alexey Dosovitskiy.
\newblock Unsupervised learning of shape and pose with differentiable point
  clouds.
\newblock In {\em Advances in neural information processing systems}, pages
  2802--2812, 2018.

\bibitem{MultiView-ji2017surfacenet}
Mengqi Ji, Juergen Gall, Haitian Zheng, Yebin Liu, and Lu~Fang.
\newblock {SurfaceNet}: An end-to-end {3D} neural network for multiview
  stereopsis.
\newblock In {\em Proceedings of the IEEE International Conference on Computer
  Vision}, pages 2307--2315, 2017.

\bibitem{ji2017surfacenet}
Mengqi Ji, Juergen Gall, Haitian Zheng, Yebin Liu, and Lu~Fang.
\newblock {SurfaceNet}: An end-to-end 3d neural network for multiview
  stereopsis.
\newblock In {\em Proceedings of the IEEE International Conference on Computer
  Vision}, pages 2307--2315, 2017.

\bibitem{JohanssonB2000-ICPR}
B.~Johansson, H.~Knutsson, and G.~Granlund.
\newblock Detecting rotational symmetries using normalized convolution.
\newblock In {\em ICPR}, pages 496--500, 2000.

\bibitem{kang2001handling}
Sing~Bing Kang, Richard Szeliski, and Jinxiang Chai.
\newblock Handling occlusions in dense multi-view stereo.
\newblock In {\em Proceedings of the 2001 IEEE Computer Society Conference on
  Computer Vision and Pattern Recognition. CVPR 2001}, volume~1, pages I--I.
  IEEE, 2001.

\bibitem{kar2017learning}
Abhishek Kar, Christian H{\"a}ne, and Jitendra Malik.
\newblock Learning a multi-view stereo machine.
\newblock In {\em Advances in neural information processing systems}, pages
  365--376, 2017.

\bibitem{kendall2017end}
Alex Kendall, Hayk Martirosyan, Saumitro Dasgupta, Peter Henry, Ryan Kennedy,
  Abraham Bachrach, and Adam Bry.
\newblock End-to-end learning of geometry and context for deep stereo
  regression.
\newblock In {\em Proceedings of the IEEE International Conference on Computer
  Vision}, pages 66--75, 2017.

\bibitem{kingma2014adam}
Diederik~P Kingma and Jimmy Ba.
\newblock Adam: A method for stochastic optimization.
\newblock {\em arXiv preprint arXiv:1412.6980}, 2014.

\bibitem{KiryatiN1998-IJCV}
N.~Kiryati and Y.~Gofman.
\newblock Detecting symmetry in grey level images: the global optimization
  approach.
\newblock {\em IJCV}, 29(1):29--45, August 1998.

\bibitem{korah2008analysis}
Thommen Korah and Christopher Rasmussen.
\newblock Analysis of building textures for reconstructing partially occluded
  facades.
\newblock In {\em European conference on computer vision}, pages 359--372.
  Springer, 2008.

\bibitem{leroy2018shape}
Vincent Leroy, Jean-S{\'e}bastien Franco, and Edmond Boyer.
\newblock Shape reconstruction using volume sweeping and learned photo
  consistency.
\newblock In {\em Proceedings of the European Conference on Computer Vision
  (ECCV)}, pages 781--796, 2018.

\bibitem{LiuY2010}
Y.~Liu, H.~Hel-Or, C.~Kaplan, and L.~van Gool.
\newblock Computational symmetry in computer vision and computer graphics.
\newblock {\em FTCGV}, 5:1--197, 2010.

\bibitem{loy2006detecting}
Gareth Loy and Jan-Olof Eklundh.
\newblock Detecting symmetry and symmetric constellations of features.
\newblock In {\em European Conference on Computer Vision}, pages 508--521.
  Springer, 2006.

\bibitem{MultiView-luo2019p}
Keyang Luo, Tao Guan, Lili Ju, Haipeng Huang, and Yawei Luo.
\newblock {P-MVSNet}: Learning patch-wise matching confidence aggregation for
  multi-view stereo.
\newblock In {\em Proceedings of the IEEE International Conference on Computer
  Vision}, pages 10452--10461, 2019.

\bibitem{MaY2003}
Y.~Ma, J.~Ko{\v{s}}eck{\'{a}}, S.~Soatto, and S.~Sastry.
\newblock {\em An {I}nvitation to 3-{D} {V}ision, {F}rom {I}mages to
  {G}eometric {M}odels}.
\newblock Springer--Verlag, New York, 2004.

\bibitem{MarolaG1989-PAMI}
G.~Marola.
\newblock On the detection of the axes of symmetry of symmetric and almost
  symmetric planar images.
\newblock {\em {PAMI}}, 11(1):104--108, January 1989.

\bibitem{mescheder2019occupancy}
Lars Mescheder, Michael Oechsle, Michael Niemeyer, Sebastian Nowozin, and
  Andreas Geiger.
\newblock Occupancy networks: Learning {3D} reconstruction in function space.
\newblock In {\em Proceedings of the IEEE Conference on Computer Vision and
  Pattern Recognition}, pages 4460--4470, 2019.

\bibitem{MundyJ1994}
J.~L. Mundy and A.~Zisserman.
\newblock Repeated structures: Image correspondence constraints and {3D}
  structure recovery.
\newblock In {\em Applications of invariance in computer vision}, pages
  89--106, 1994.

\bibitem{newell2016stacked}
Alejandro Newell, Kaiyu Yang, and Jia Deng.
\newblock Stacked hourglass networks for human pose estimation.
\newblock In {\em European conference on computer vision}, pages 483--499.
  Springer, 2016.

\bibitem{park2019deepsdf}
Jeong~Joon Park, Peter Florence, Julian Straub, Richard Newcombe, and Steven
  Lovegrove.
\newblock {DeepSDF}: Learning continuous signed distance functions for shape
  representation.
\newblock {\em arXiv preprint arXiv:1901.05103}, 2019.

\bibitem{schonberger2016pixelwise}
Johannes~L Sch{\"o}nberger, Enliang Zheng, Jan-Michael Frahm, and Marc
  Pollefeys.
\newblock Pixelwise view selection for unstructured multi-view stereo.
\newblock In {\em European Conference on Computer Vision}, pages 501--518.
  Springer, 2016.

\bibitem{Equivalent2ImageClassification}
Maxim Tatarchenko, Stephan~R Richter, Ren{\'e} Ranftl, Zhuwen Li, Vladlen
  Koltun, and Thomas Brox.
\newblock What do single-view {3D} reconstruction networks learn?
\newblock In {\em Proceedings of the IEEE Conference on Computer Vision and
  Pattern Recognition}, pages 3405--3414, 2019.

\bibitem{tola2012efficient}
Engin Tola, Christoph Strecha, and Pascal Fua.
\newblock Efficient large-scale multi-view stereo for ultra high-resolution
  image sets.
\newblock {\em Machine Vision and Applications}, 23(5):903--920, 2012.

\bibitem{Troje-symmetry}
N.F. Troje and H.H. Bulthoff.
\newblock How is bilateral symmetry of human faces used for recognition of
  novel views.
\newblock {\em Vision Research}, 38(1):79--89, 1998.

\bibitem{Uhrig2017THREEDV}
Jonas Uhrig, Nick Schneider, Lukas Schneider, Uwe Franke, Thomas Brox, and
  Andreas Geiger.
\newblock Sparsity invariant cnns.
\newblock In {\em International Conference on 3D Vision (3DV)}, 2017.

\bibitem{Vetter-symmetry2}
T.~Vetter, T.~Poggio, and H.~H. Bulthoff.
\newblock The importance of symmetry and virtual views in three-dimensional
  object recognition.
\newblock {\em Current Biology}, 4:18--23, 1994.

\bibitem{wang2018pixel2mesh}
Nanyang Wang, Yinda Zhang, Zhuwen Li, Yanwei Fu, Wei Liu, and Yu-Gang Jiang.
\newblock {Pixel2mesh}: Generating {3D} mesh models from single {RGB} images.
\newblock In {\em Proceedings of the European Conference on Computer Vision
  (ECCV)}, pages 52--67, 2018.

\bibitem{wu2018learning}
Jiajun Wu, Chengkai Zhang, Xiuming Zhang, Zhoutong Zhang, William~T Freeman,
  and Joshua~B Tenenbaum.
\newblock Learning shape priors for single-view {3D} completion and
  reconstruction.
\newblock In {\em Proceedings of the European Conference on Computer Vision
  (ECCV)}, pages 646--662, 2018.

\bibitem{wu2019unsupervised}
Shangzhe Wu, Christian Rupprecht, and Andrea Vedaldi.
\newblock Unsupervised learning of probably symmetric deformable 3d objects
  from images in the wild.
\newblock In {\em Conference on Computer Vision and Pattern Recognition}, 2020.

\bibitem{xu2019disn}
Qiangeng Xu, Weiyue Wang, Duygu Ceylan, Radomir Mech, and Ulrich Neumann.
\newblock {DISN}: Deep implicit surface network for high-quality single-view 3d
  reconstruction.
\newblock In {\em Advances in Neural Information Processing Systems}, pages
  490--500, 2019.

\bibitem{MultiView-xue2019mvscrf}
Youze Xue, Jiansheng Chen, Weitao Wan, Yiqing Huang, Cheng Yu, Tianpeng Li, and
  Jiayu Bao.
\newblock {MVSCRF}: Learning multi-view stereo with conditional random fields.
\newblock In {\em Proceedings of the IEEE International Conference on Computer
  Vision}, pages 4312--4321, 2019.

\bibitem{yan2016perspective}
Xinchen Yan, Jimei Yang, Ersin Yumer, Yijie Guo, and Honglak Lee.
\newblock Perspective transformer nets: Learning single-view {3D} object
  reconstruction without {3D} supervision.
\newblock In {\em Advances in Neural Information Processing Systems}, pages
  1696--1704, 2016.

\bibitem{yao2018mvsnet}
Yao Yao, Zixin Luo, Shiwei Li, Tian Fang, and Long Quan.
\newblock {MVSNet}: Depth inference for unstructured multi-view stereo.
\newblock In {\em Proceedings of the European Conference on Computer Vision
  (ECCV)}, pages 767--783, 2018.

\bibitem{ZabrodskyH1995-PAMI}
H.~Zabrodsky, S.~Peleg, and D.~Avnir.
\newblock Symmetry as a continuous feature.
\newblock {\em PAMI}, 17(12):1154--1166, December 1995.

\bibitem{zbontar2016stereo}
Jure Zbontar, Yann LeCun, et~al.
\newblock Stereo matching by training a convolutional neural network to compare
  image patches.
\newblock {\em Journal of Machine Learning Research}, 17(1-32):2, 2016.

\bibitem{zhang2020portrait}
Xuaner~Cecilia Zhang, Yun-Ta Tsai, Rohit Pandey, Xiuming Zhang, Ren Ng, David~E
  Jacobs, et~al.
\newblock Portrait shadow manipulation.
\newblock {\em arXiv preprint arXiv:2005.08925}, 2020.

\bibitem{zhou2019continuity}
Yi~Zhou, Connelly Barnes, Jingwan Lu, Jimei Yang, and Hao Li.
\newblock On the continuity of rotation representations in neural networks.
\newblock In {\em Proceedings of the IEEE Conference on Computer Vision and
  Pattern Recognition}, pages 5745--5753, 2019.

\bibitem{zhou2019neurvps}
Yichao Zhou, Haozhi Qi, Jingwei Huang, and Yi~Ma.
\newblock Neurvps: Neural vanishing point scanning via conic convolution.
\newblock In {\em Advances in Neural Information Processing Systems}, pages
  864--873, 2019.

\end{thebibliography}

\clearpage
\appendix
\section{Supplementary Materials}
\subsection{Derivation of the Relationship between $\C$ and $\w$}
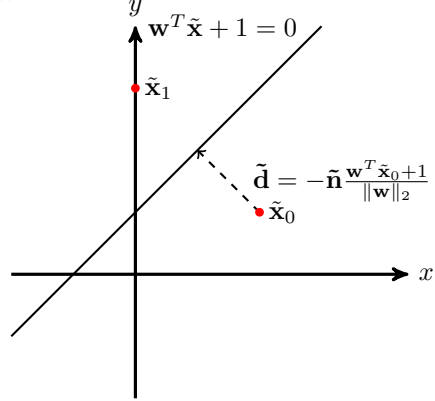
\begin{wrapfigure}{r}{0.44\textwidth}
\centering
\vspace{-25pt}
\begin{tikzpicture}[
    scale=3.3,
    axis/.style={very thick, ->, >=stealth'},
    important line/.style={thick},
    every node/.style={color=black}
    ]
    \draw[axis] (-0.5,0)  -- (1.1,0) node(xline)[right] {$x$};
    \draw[axis] (0,-0.5) -- (0,1.0) node(yline)[above] {$y$};
    \draw[important line] (.-0.5,-0.25) coordinate (A) -- (.75,1)
        coordinate (B) node[left] {$\w^T\tilde\x+1=0\;\;$};
    \coordinate (x0) at (.50,.25); 
    \coordinate (x1) at (.00,.75); 
    \coordinate (xm) at (.25,.50); 
    \draw[important line,dashed,->] (x0) -- node[right] {$\;\;\mathbf{\vec{d}}=-\mathbf{\vec{n}} \frac{\w^T\tilde\x_0+1}{\|\w\|_2}$} (xm);
    \fill[red] (x0) circle (.5pt) node[right] {$\tilde\x_0$};
    \fill[red] (x1) circle (.5pt) node[right] {$\tilde\x_1$};
\end{tikzpicture}
\caption{Mirroring a point.}
\label{fig:mirroring}
\end{wrapfigure}

Let $\tilde\x_0 \in \mathbb{R}^3$ be a point in the camera space and $\tilde\x_1$ be its mirror point with respect to the symmetry plane
\begin{equation}
    \w^T \tilde \x + 1 = 0.
\end{equation}
As shown in \Cref{fig:mirroring}, the normal of the plane is $\mathbf{\vec{n}}=\frac{\w}{\|\w\|_2}$ and 
the distance between $\tilde\x_0$ and the symmetry plane is 
$\frac{\w^T\tilde\x_0+1}{\|\w\|_2}$ using the formula of distance from a point to a plane.  Therefore, we have
\begin{equation}
    \tilde\x_1 = \tilde\x_0 - 2\frac{\w^T\tilde\x_0+1}{\|\w\|_2^2}\w.
\end{equation}
We could also write this in matrix form:
\begin{equation}
    \begin{bmatrix} \tilde\x_1 \\ 1 \end{bmatrix} =
    \begin{bmatrix} \tilde\x_0 \\ 1 \end{bmatrix} \begin{bmatrix}
      \mathbf{I} - \frac{2\w\w^T}{\|\w\|_2^2} & -\frac{2\w}{\|\w\|_2^2} \\
      \mathbf{0} & 1 \\
    \end{bmatrix}.
\end{equation}

\begin{wrapfigure}{r}{0.44\textwidth}
\centering
\includegraphics[width=.95\linewidth]{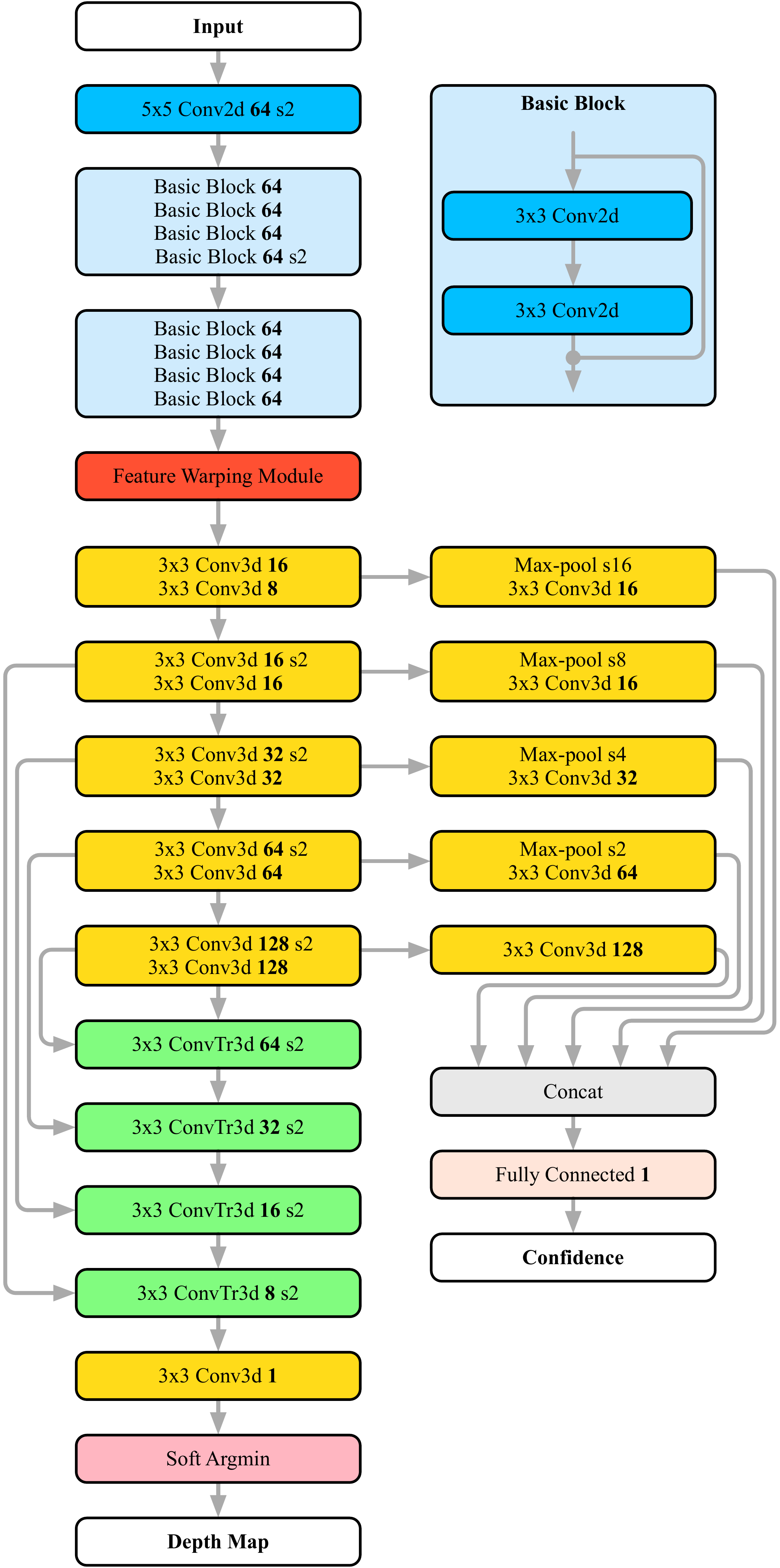}
\caption{Illustration of SymmetryNet's network architecture.  We show the resulting number of channels after each operator in \textbf{bold}. ``s2'' represents stride-2 operators.}
\vspace{2mm}
\label{fig:architecture}
\vspace{-17mm}
\end{wrapfigure}

Because the transformation between the camera space and the pixel space is given by
\begin{equation}
    \x = \K \begin{bmatrix} \tilde\x \\ 1 \end{bmatrix},
\end{equation}
we finally have
\begin{align*}
  \C &= \K \begin{bmatrix}
  \mathbf{I} - \frac{2\w\w^T}{\|\w\|_2^2} & -\frac{2\w}{\|\w\|_2^2} \\
  \mathbf{0} & 1 \\
  \end{bmatrix}\K^{-1} \\
  &= \K \left(\mathbf{I} - \frac{2}{\|\w\|_2^2}\begin{bmatrix}
  \w \\
  \mathbf{0} \\
  \end{bmatrix}
  \begin{bmatrix} \w^T& \mathbf{1} \end{bmatrix} \right)\K^{-1}.
\end{align*}

\subsection{Network Architecture}

We show the diagram of SymmetryNet's network architecture in \Cref{fig:architecture}.

\subsection{Learning with More Symmetries}  \label{sec:multiaxis}
\begin{table*}[tpb]
\centering
\renewcommand{\arraystretch}{1.2}
\setlength{\tabcolsep}{1.2mm}

\resizebox{\textwidth}{!}{%
\begin{tabular}{l|cccc|cccc|cccc}
\hline
\multirow{2}{*}{}   & \multicolumn{4}{c|}{Mean $\ell_1$}                                                                                            & \multicolumn{4}{c|}{Median $\ell_1$}                                                                                          & \multicolumn{4}{c}{RMSE}                                                                                                      \\ \cline{2-13} 
                    & \scriptsize $|\mathcal{M}|=1$ & \scriptsize $|\mathcal{M}|=2$ & \scriptsize $|\mathcal{M}|=3$ & \scriptsize $|\mathcal{M}|=4$ & \scriptsize $|\mathcal{M}|=1$ & \scriptsize $|\mathcal{M}|=2$ & \scriptsize $|\mathcal{M}|=3$ & \scriptsize $|\mathcal{M}|=4$ & \scriptsize $|\mathcal{M}|=1$ & \scriptsize $|\mathcal{M}|=2$ & \scriptsize $|\mathcal{M}|=3$ & \scriptsize $|\mathcal{M}|=4$ \\ \hline
Plane               & 0.0278                        & 0.0146                        & 0.0154                        & \textbf{0.0082}               & 0.0190                        & 0.0075                        & 0.0080                        & \textbf{0.0039}               & 0.0407                        & 0.0256                        & 0.0264                        & \textbf{0.0153}               \\
Bench               & 0.0211                        & \textbf{0.0104}               & 0.0139                        & 0.0109                        & 0.0120                        & \textbf{0.0037}               & 0.0065                        & 0.0054                        & 0.0355                        & 0.0240                        & 0.0272                        & \textbf{0.0215}               \\
Cabinet$^\sharp$    & 0.0252                        & 0.0111                        & 0.0111                        & \textbf{0.0087}               & 0.0145                        & 0.0059                        & 0.0055                        & \textbf{0.0030}               & 0.0415                        & \textbf{0.0206}               & 0.0219                        & 0.0211                        \\
Car                 & 0.0196                        & 0.0113                        & \textbf{0.0094}               & 0.0136                        & 0.0125                        & 0.0056                        & \textbf{0.0032}               & 0.0063                        & 0.0311                        & \textbf{0.0220}               & 0.0224                        & 0.0265                        \\
Chair               & 0.0215                        & \textbf{0.0101}               & 0.0111                        & 0.0176                        & 0.0121                        & \textbf{0.0050}               & 0.0059                        & 0.0077                        & 0.0368                        & \textbf{0.0191}               & 0.0202                        & 0.0316                        \\
Monitor$^\sharp$    & 0.0323                        & 0.0125                        & \textbf{0.0084}               & 0.0123                        & 0.0214                        & 0.0063                        & \textbf{0.0040}               & 0.0065                        & 0.0474                        & 0.0239                        & \textbf{0.0152}               & 0.0233                        \\
Lamp                & 0.0167                        & \textbf{0.0085}               & 0.0139                        & 0.0137                        & 0.0109                        & \textbf{0.0043}               & 0.0073                        & 0.0073                        & 0.0261                        & \textbf{0.0171}               & 0.0240                        & 0.0235                        \\
Speaker$^\sharp$    & 0.0304                        & 0.
0153                        & \textbf{0.0085}               & 0.0139                        & 0.0210                        & 0.0073                        & \textbf{0.0045}               & 0.0065                        & 0.0441                        & 0.0279                        & \textbf{0.0168}               & 0.0252                        \\
Firearm             & 0.0289                        & \textbf{0.0140}               & 0.0186                        & 0.0154                        & 0.0184                        & \textbf{0.0064}               & 0.0081                        & 0.0080                        & 0.0447                        & 0.0272                        & 0.0336                        & \textbf{0.0264}               \\
Couch               & 0.0251                        & \textbf{0.0086}               & 0.0145                        & 0.0111                        & 0.0163                        & \textbf{0.0040}               & 0.0069                        & 0.0059                        & 0.0372                        & \textbf{0.0157}               & 0.0267                        & 0.0199                        \\
Table$^\sharp$      & 0.0255                        & 0.0148                        & 0.0127                        & \textbf{0.0079}               & 0.0150                        & 0.0071                        & 0.0067                        & \textbf{0.0040}               & 0.0399                        & 0.0263                        & 0.0241                        & \textbf{0.0163}               \\
Phone$^\sharp$      & 0.0333                        & 0.0158                        & \textbf{0.0104}               & 0.0137                        & 0.0237                        & 0.0079                        & \textbf{0.0051}               & 0.0063                        & 0.0463                        & 0.0279                        & \textbf{0.0196}               & 0.0252                        \\
Watercraft$^\sharp$ & 0.0361                        & 0.0190                        & 0.0144                        & \textbf{0.0097}               & 0.0216                        & 0.0084                        & 0.0069                        & \textbf{0.0049}               & 0.0547                        & 0.0337                        & 0.0258                        & \textbf{0.0184}               \\ \hline
Average             & 0.0263                        & 0.0127                        & 0.0125                        & \textbf{0.0120}               & 0.0162                        & 0.0059                        & 0.0059                        & \textbf{0.0055}               & 0.0410                        & 0.0242                        & 0.0236                        & \textbf{0.0229}               \\ \hline
\end{tabular}%
}
\caption{Performance of SymmetryNet with multiple symmetry transformations. Let $\mathcal{M}$ be the set of symmetry transformations that the objects admit, including the identity $\M_1=\I$. When $|\mathcal{M}|=1$, we concatenate the image feature with the one-hot features on the depth dimension to construct the cost volume, while for $|\mathcal{M}|=2$, $|\mathcal{M}|=3$, $|\mathcal{M}|=4$, we gradually add the warped features to the cost volume from the transformations $\M_2=\mathrm{diag}(-1, 1, 1)$, $\M_3=\mathrm{diag}(1, -1, 1)$, and $\M_4=\mathrm{diag}(-1, -1, 1)$.  We label the categories with $\sharp$ if both of the mean $\ell_1$ errors in the columns of $|\mathcal{M}|=3$ and $|\mathcal{M}|=4$ are smaller than the error in the column of $|\mathcal{M}|=2$, indicating many objects in these categories admit additional symmetry.  The lowest error in each comparison is highlighted in \textbf{boldface}.}
\label{tab:multiaxis}

\vspace{5mm}

\end{table*}

The ShapeNet dataset only detects one of the bilateral symmetry planes and aligns each model to the Y-Z plane.  However, many objects in the dataset admit more than one symmetries. For example, a centered rectangle admits four different symmetry transformations, i.e., $\M_1=\I$, $\M_2 = \diag(-1, 1, 1)$, $\M_3 = \diag(1, -1, 1)$, and $\M_4 = \diag(-1, -1, 1)$ for the $\M$ in \Cref{eq:correspondence}.  Therefore, one may ask whether those extra symmetries could help the depth estimation.  To answer this question, we conduct 4 additional experiments.  In each experiment, we assume that the ground truth camera pose $\Rt$ is known, treat $\{\C_n=\K\Rt\M_n\Rt^{-1}\K^{-1}\}_{n=1}^4$ with respect to each symmetry transformation as input, and only evaluate the depth branch of SymmetryNet. For the $i$th experiment, we use the symmetry transformation $\{\C_n\}_{n=1}^i$ as the input to  gradually introduce more symmetry transformations into the feature warping module.  For the case $i=1$, we also concatenate a one-hot vector representing the depth to the cost volume to prevent all the feature being the same along the depth dimension.

The results are shown in \Cref{tab:multiaxis}.  We find that performance gain is significant after introducing the main bilateral symmetry $\M_2$, as it makes the network geometry-aware for the first time.  Adding $\M_3$ and $\M_4$ gives barely any performance improvement overall.  This is probably because the symmetry transformations $\M_3$ and $\M_4$ might not be accurate, as the second bilateral symmetry plane is not explicitly aligned to the X-Z plane in the dataset. However, there is an interesting phenomenon that catches our eyes:  For categories marked with \#, such as cabinets, monitors, speakers, tables, phones, and water crafts that often admit additional bilateral symmetry, adding more transformations to the warping module almost always improves the performance, as they help provide additional information in case that the main symmetry is compromised due to occlusion. For categories such as benches, chairs, firearms, and couches that are unlikely to have a second bilateral symmetry, using additional warped features degrades the performance. We believe that this is mainly resulting from the inconsistency and noise introduced by the new features. 

\subsection{Failure Cases}
\begin{figure}[tbp]
  \centering
  \setlength{\lineskip}{0pt}
  \includegraphics[width=\linewidth]{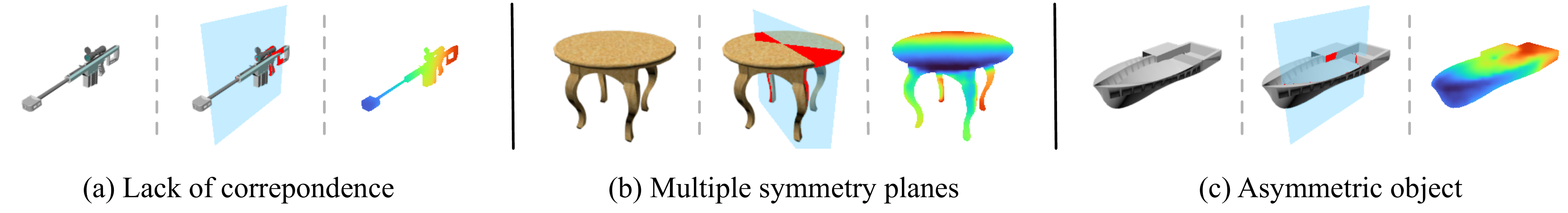}
  \caption{Sampled failure cases of SymmetryNet on ShapeNet.}
  \label{fig:failure}
\end{figure}

\Cref{fig:failure} shows sampled failure cases on ShapeNet.  We categorize those cases into three classes: lack of correspondence, existence of multiple symmetry planes, and asymetric objects.  For the first category, e.g., the firearm shown in \Cref{fig:failure}(a), it is hard to accurately find the symmetry plane from the geometry cues because for most pixels, their corresponding points are occluded and invisible in the picture.  For the second category, objects in shapes such as squares and cylinders admit multiple reflection symmetry, and SymmetryNet may return the reflection plane that differs from the symmetry plane of the ground truth.  For the third category, some objects in ShapeNet are not symmetric.  Thus, the detected symmetry plane might be different from the ``ground truth symmetry plane'' computed by applying $\Rt$ to the Y-Z plane in the world space.

\subsection{More Result Visualization}

\begin{figure}[h]
  \centering
  \setlength{\lineskip}{0pt}
  \includegraphics[width=0.11\linewidth]{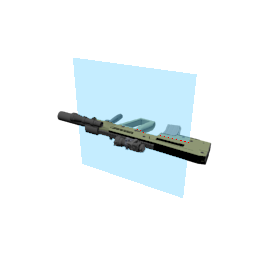}%
  \includegraphics[width=0.11\linewidth]{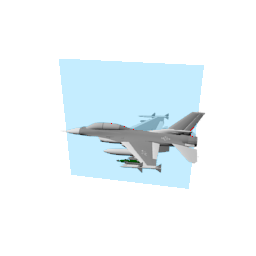}%
  \includegraphics[width=0.11\linewidth]{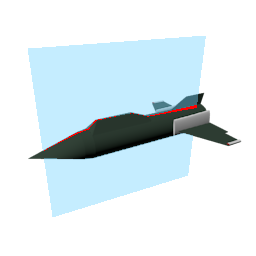}%
  \includegraphics[width=0.11\linewidth]{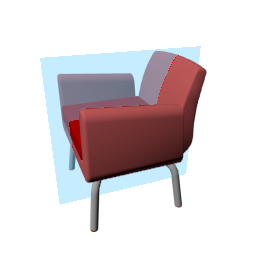}%
  \includegraphics[width=0.11\linewidth]{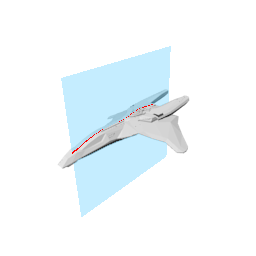}%
  \includegraphics[width=0.11\linewidth]{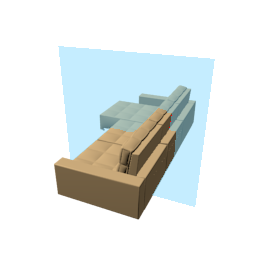}%
  \includegraphics[width=0.11\linewidth]{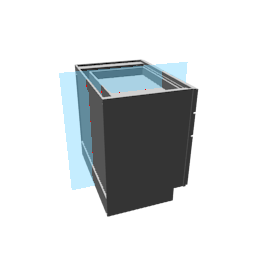}%
  \includegraphics[width=0.11\linewidth]{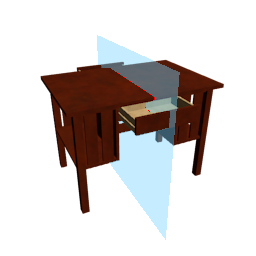}%
  \includegraphics[width=0.11\linewidth]{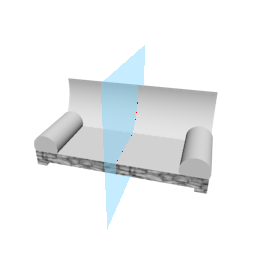}%

  \includegraphics[width=0.11\linewidth]{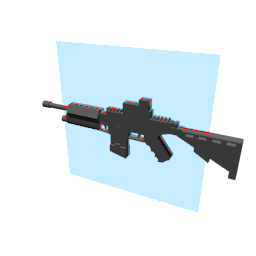}%
  \includegraphics[width=0.11\linewidth]{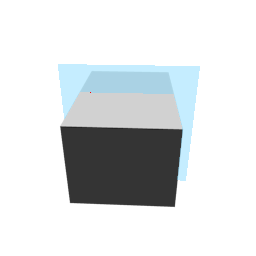}%
  \includegraphics[width=0.11\linewidth]{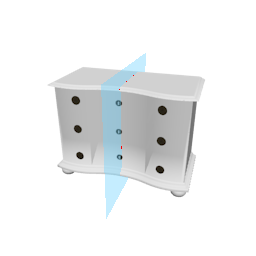}%
  \includegraphics[width=0.11\linewidth]{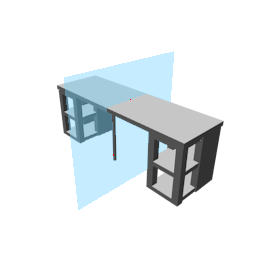}%
  \includegraphics[width=0.11\linewidth]{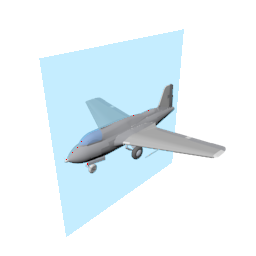}%
  \includegraphics[width=0.11\linewidth]{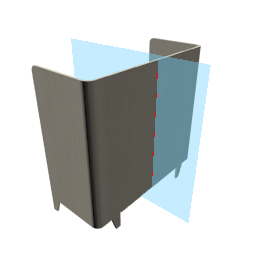}%
  \includegraphics[width=0.11\linewidth]{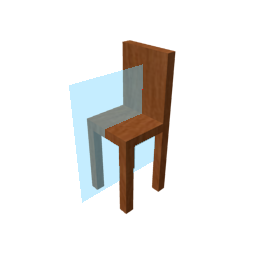}%
  \includegraphics[width=0.11\linewidth]{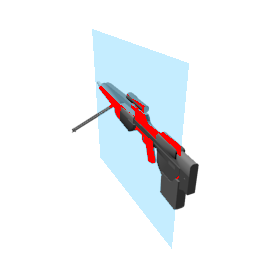}%
  \includegraphics[width=0.11\linewidth]{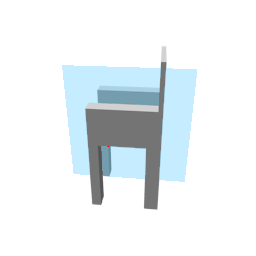}%

  \includegraphics[width=0.11\linewidth]{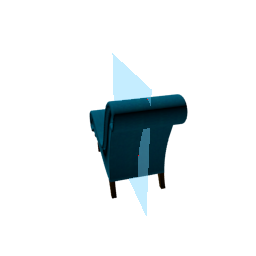}%
  \includegraphics[width=0.11\linewidth]{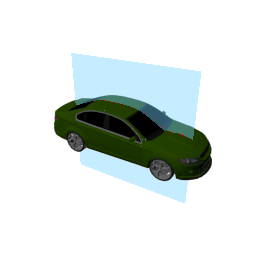}%
  \includegraphics[width=0.11\linewidth]{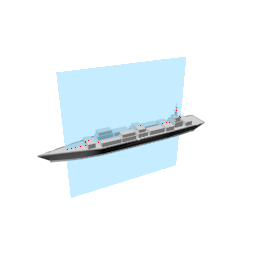}%
  \includegraphics[width=0.11\linewidth]{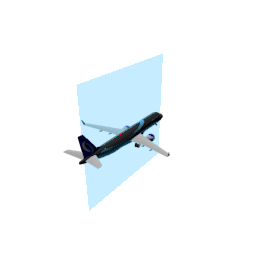}%
  \includegraphics[width=0.11\linewidth]{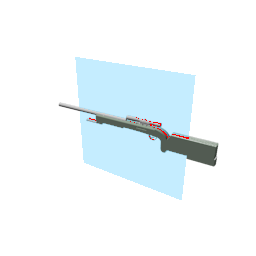}%
  \includegraphics[width=0.11\linewidth]{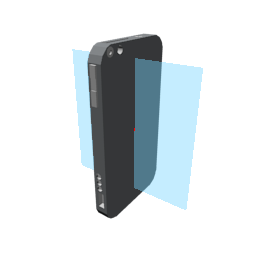}%
  \includegraphics[width=0.11\linewidth]{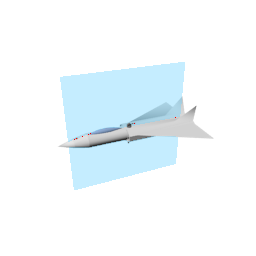}%
  \includegraphics[width=0.11\linewidth]{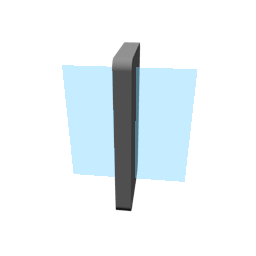}%
  \includegraphics[width=0.11\linewidth]{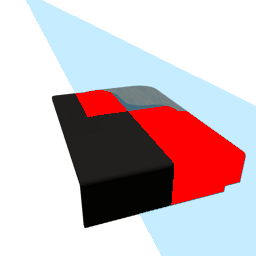}%

  \includegraphics[width=0.11\linewidth]{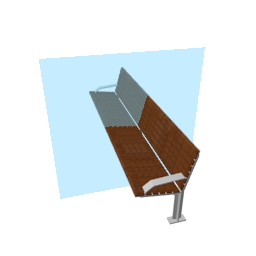}%
  \includegraphics[width=0.11\linewidth]{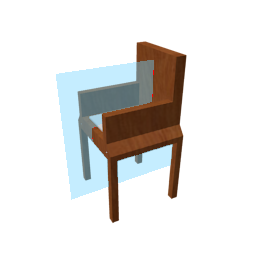}%
  \includegraphics[width=0.11\linewidth]{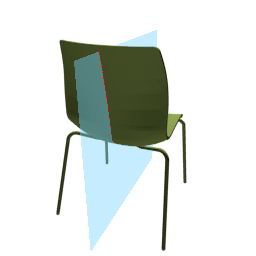}%
  \includegraphics[width=0.11\linewidth]{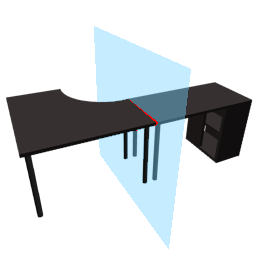}%
  \includegraphics[width=0.11\linewidth]{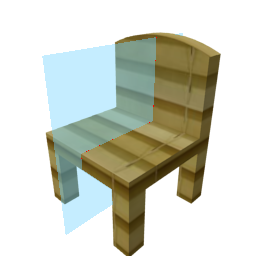}%
  \includegraphics[width=0.11\linewidth]{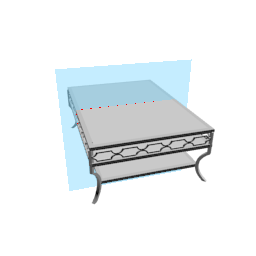}%
  \includegraphics[width=0.11\linewidth]{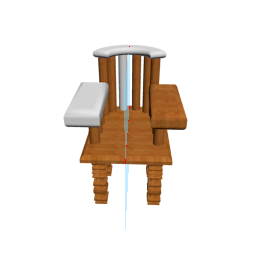}%
  \includegraphics[width=0.11\linewidth]{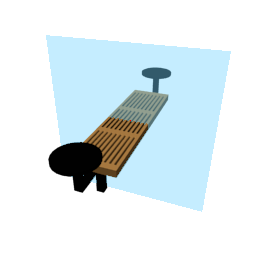}%
  \includegraphics[width=0.11\linewidth]{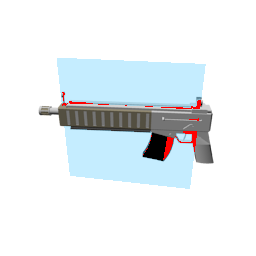}%

  \includegraphics[width=0.11\linewidth]{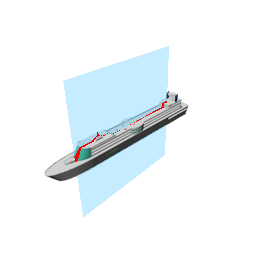}%
  \includegraphics[width=0.11\linewidth]{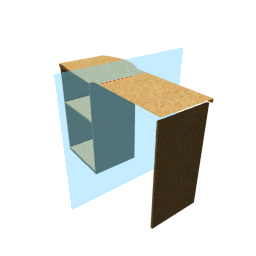}%
  \includegraphics[width=0.11\linewidth]{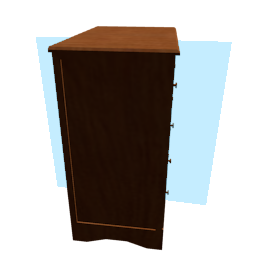}%
  \includegraphics[width=0.11\linewidth]{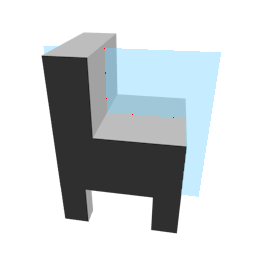}%
  \includegraphics[width=0.11\linewidth]{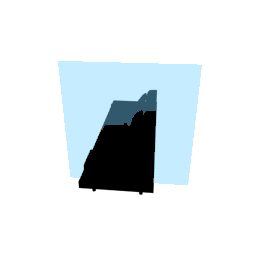}%
  \includegraphics[width=0.11\linewidth]{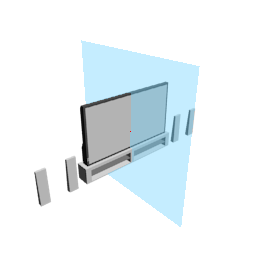}%
  \includegraphics[width=0.11\linewidth]{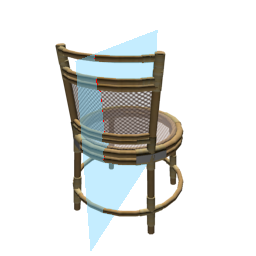}%
  \includegraphics[width=0.11\linewidth]{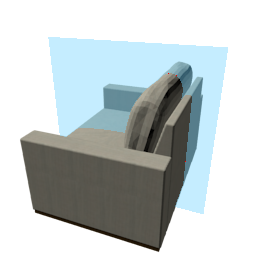}%
  \includegraphics[width=0.11\linewidth]{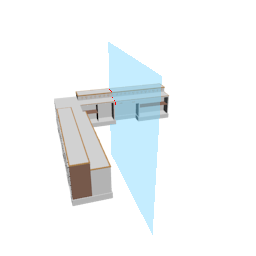}%

  \includegraphics[width=0.11\linewidth]{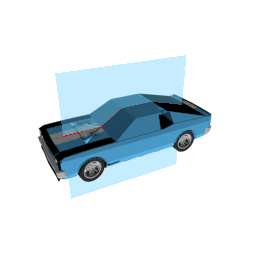}%
  \includegraphics[width=0.11\linewidth]{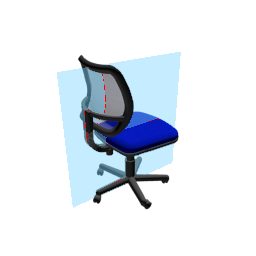}%
  \includegraphics[width=0.11\linewidth]{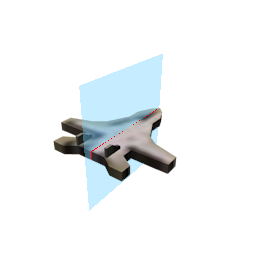}%
  \includegraphics[width=0.11\linewidth]{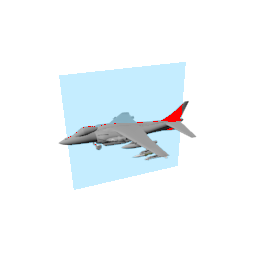}%
  \includegraphics[width=0.11\linewidth]{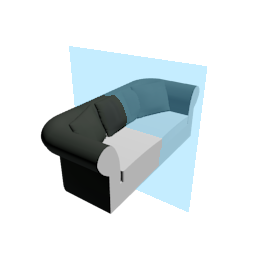}%
  \includegraphics[width=0.11\linewidth]{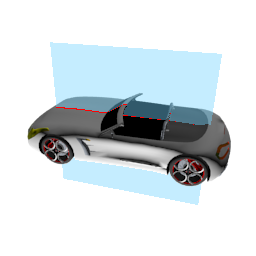}%
  \includegraphics[width=0.11\linewidth]{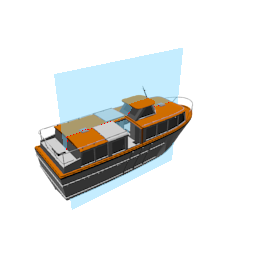}%
  \includegraphics[width=0.11\linewidth]{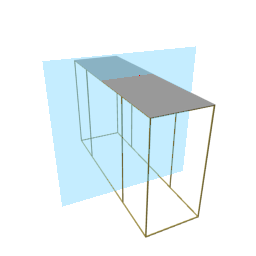}%
  \includegraphics[width=0.11\linewidth]{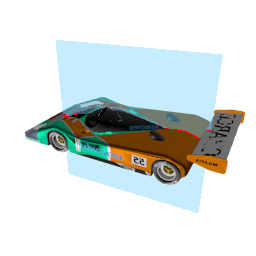}%

  \includegraphics[width=0.11\linewidth]{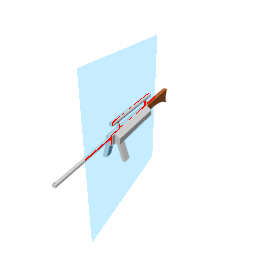}%
  \includegraphics[width=0.11\linewidth]{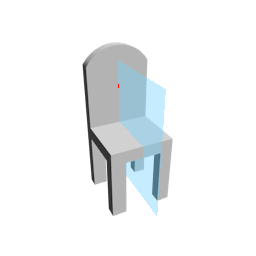}%
  \includegraphics[width=0.11\linewidth]{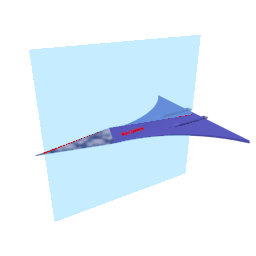}%
  \includegraphics[width=0.11\linewidth]{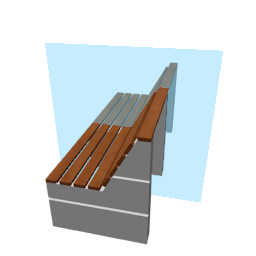}%
  \includegraphics[width=0.11\linewidth]{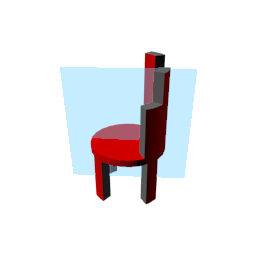}%
  \includegraphics[width=0.11\linewidth]{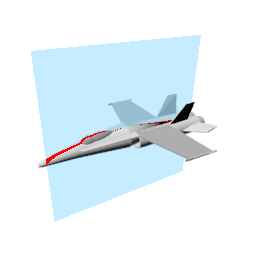}%
  \includegraphics[width=0.11\linewidth]{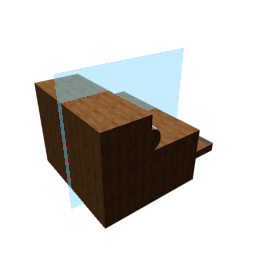}%
  \includegraphics[width=0.11\linewidth]{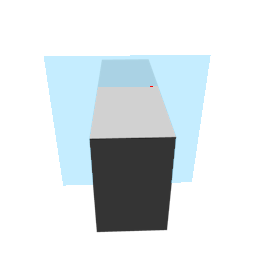}%
  \includegraphics[width=0.11\linewidth]{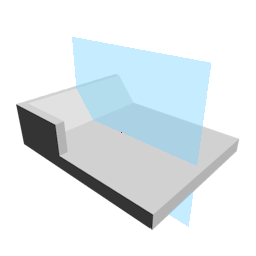}%

  \includegraphics[width=0.11\linewidth]{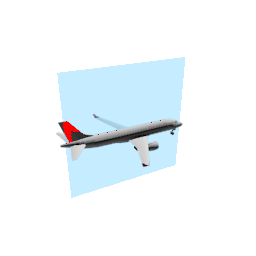}%
  \includegraphics[width=0.11\linewidth]{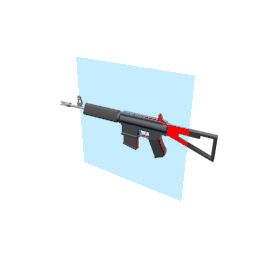}%
  \includegraphics[width=0.11\linewidth]{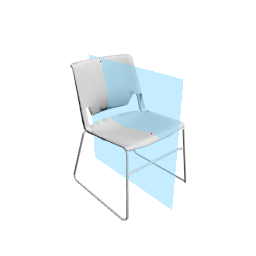}%
  \includegraphics[width=0.11\linewidth]{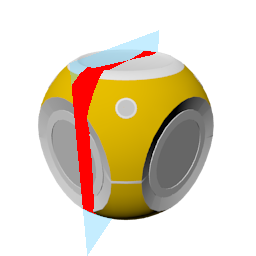}%
  \includegraphics[width=0.11\linewidth]{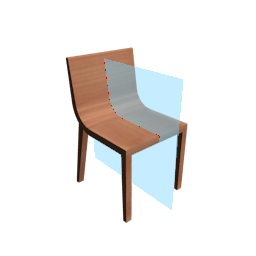}%
  \includegraphics[width=0.11\linewidth]{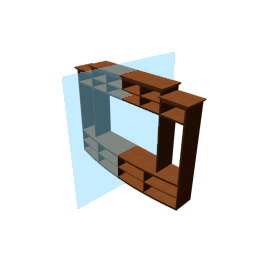}%
  \includegraphics[width=0.11\linewidth]{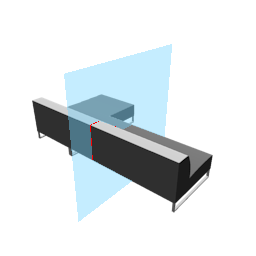}%
  \includegraphics[width=0.11\linewidth]{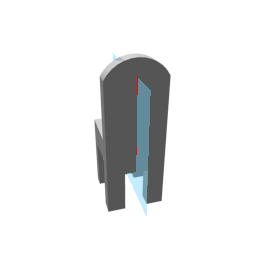}%
  \includegraphics[width=0.11\linewidth]{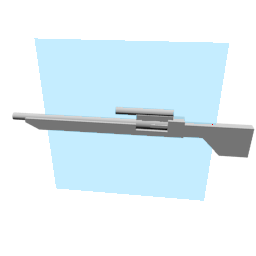}%

  \includegraphics[width=0.11\linewidth]{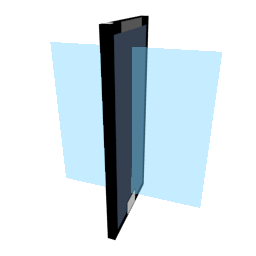}%
  \includegraphics[width=0.11\linewidth]{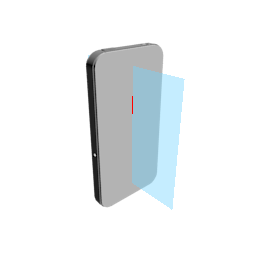}%
  \includegraphics[width=0.11\linewidth]{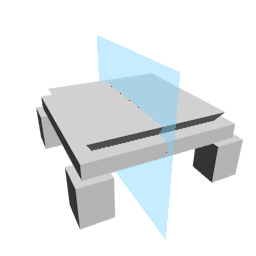}%
  \includegraphics[width=0.11\linewidth]{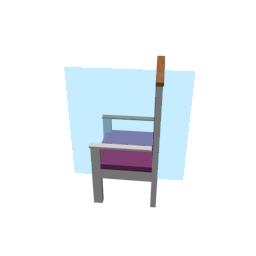}%
  \includegraphics[width=0.11\linewidth]{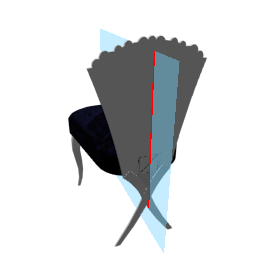}%
  \includegraphics[width=0.11\linewidth]{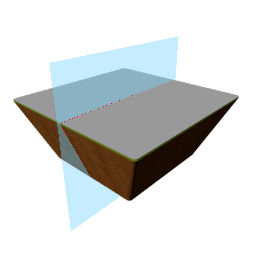}%
  \includegraphics[width=0.11\linewidth]{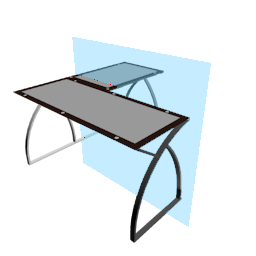}%
  \includegraphics[width=0.11\linewidth]{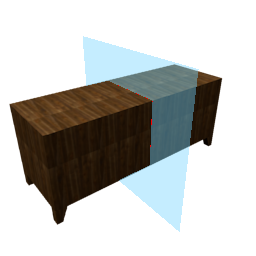}%
  \includegraphics[width=0.11\linewidth]{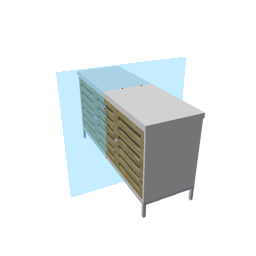}%

  \includegraphics[width=0.11\linewidth]{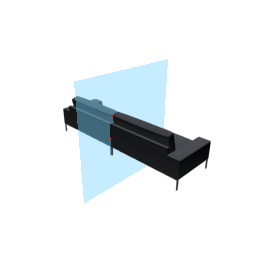}%
  \includegraphics[width=0.11\linewidth]{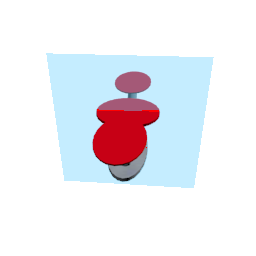}%
  \includegraphics[width=0.11\linewidth]{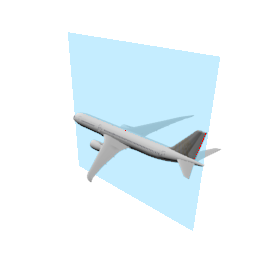}%
  \includegraphics[width=0.11\linewidth]{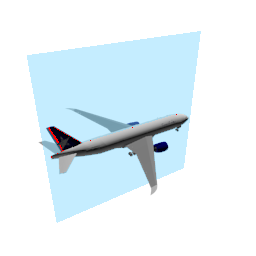}%
  \includegraphics[width=0.11\linewidth]{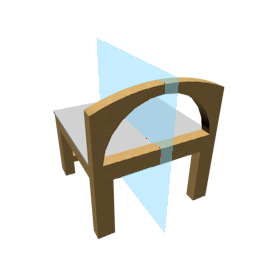}%
  \includegraphics[width=0.11\linewidth]{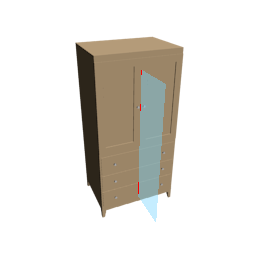}%
  \includegraphics[width=0.11\linewidth]{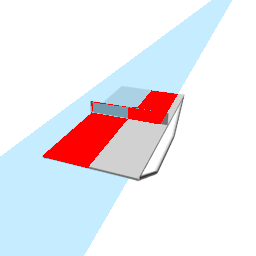}%
  \includegraphics[width=0.11\linewidth]{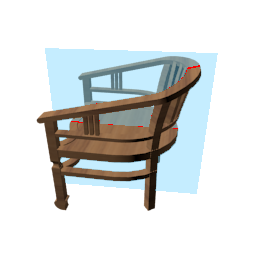}%
  \includegraphics[width=0.11\linewidth]{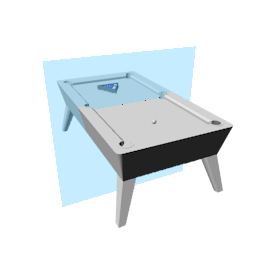}%

  \includegraphics[width=0.11\linewidth]{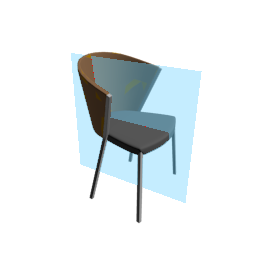}%
  \includegraphics[width=0.11\linewidth]{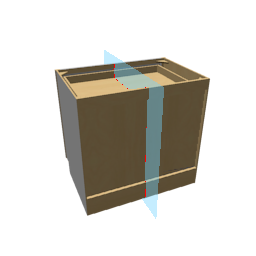}%
  \includegraphics[width=0.11\linewidth]{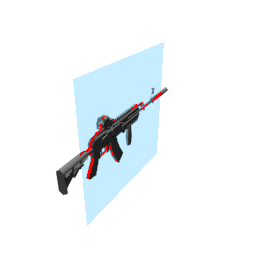}%
  \includegraphics[width=0.11\linewidth]{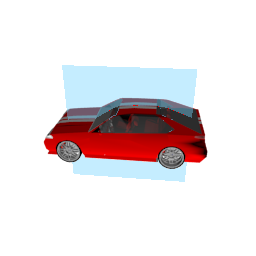}%
  \includegraphics[width=0.11\linewidth]{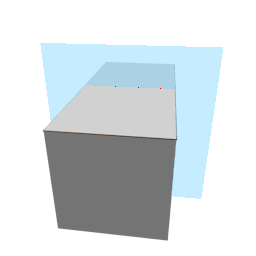}%
  \includegraphics[width=0.11\linewidth]{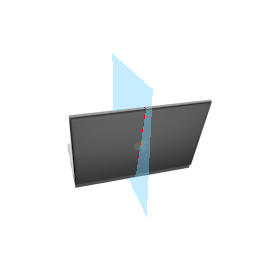}%
  \includegraphics[width=0.11\linewidth]{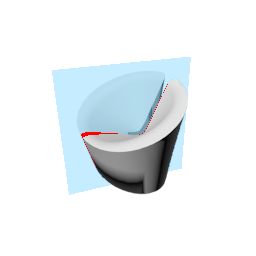}%
  \includegraphics[width=0.11\linewidth]{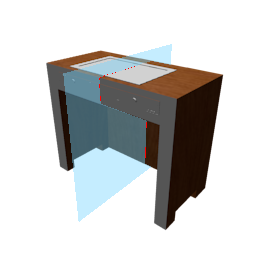}%
  \includegraphics[width=0.11\linewidth]{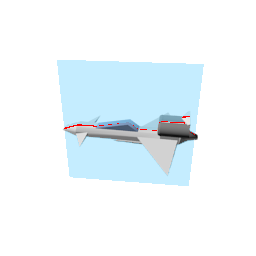}%

  \includegraphics[width=0.11\linewidth]{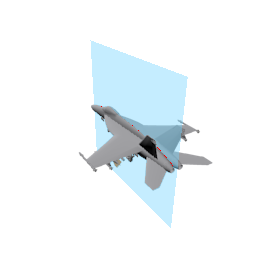}%
  \includegraphics[width=0.11\linewidth]{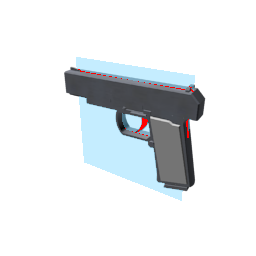}%
  \includegraphics[width=0.11\linewidth]{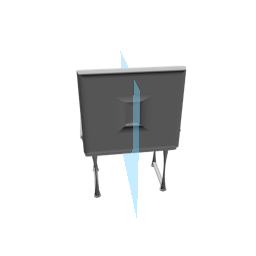}%
  \includegraphics[width=0.11\linewidth]{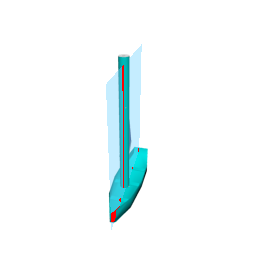}%
  \includegraphics[width=0.11\linewidth]{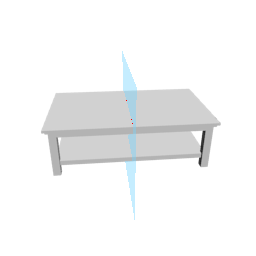}%
  \includegraphics[width=0.11\linewidth]{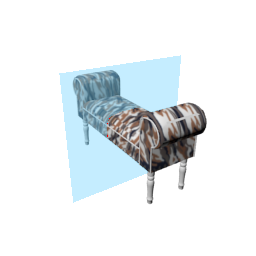}%
  \includegraphics[width=0.11\linewidth]{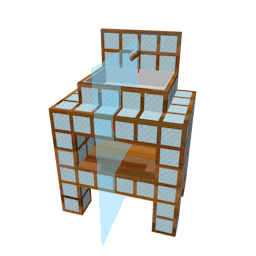}%
  \includegraphics[width=0.11\linewidth]{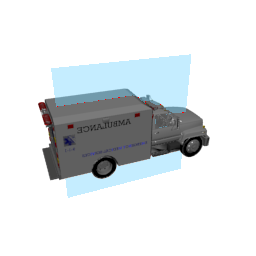}%
  \includegraphics[width=0.11\linewidth]{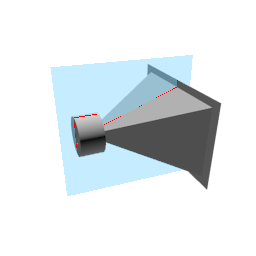}%

  \caption{Detected symmetry planes of SymmetryNet on \emph{random sampled images} from ShapeNet.  Errors of symmetry planes, i.e., pixels between the predicted plane and the ground truth plane, are highlighted in red.}
  \label{fig:random-planes}
\end{figure}

\Cref{fig:random-planes} shows the visual quality of the detected symmetry planes of SymmetryNet  on \textbf{random  sampled} testing images from ShapeNet. We find that for most images, SymmetryNet is able to determine the normal of symmetry plane accurately by utilization the geometric constraints from symmetry.  The errors (red pixels) are sub-pixel in most cases.

\end{document}